\documentclass{article}

\usepackage[preprint,nonatbib]{neurips_2020}

\usepackage[utf8]{inputenc} 
\usepackage[T1]{fontenc}    
\usepackage{hyperref}       
\usepackage{url}            
\usepackage{booktabs}       
\usepackage{amsfonts}       
\usepackage{nicefrac}       
\usepackage{microtype}      
\usepackage{graphicx}
\usepackage{amsmath}
\usepackage{mathtools}
\usepackage{xcolor}
\usepackage{listings}
\usepackage{sidecap}
\usepackage{slashbox}

\usepackage{amsmath}
\usepackage{amsfonts}
\usepackage{bm}

\def\eqref#1{equation~\ref{#1}}

\def\1{\bm{1}}

\DeclareMathAlphabet{\mathsfit}{\encodingdefault}{\sfdefault}{m}{sl}
\SetMathAlphabet{\mathsfit}{bold}{\encodingdefault}{\sfdefault}{bx}{n}

\title{Analysis of Random Perturbations for Robust Convolutional Neural Networks}

\author{%
  Adam Dziedzic, Sanjay Krishnan\\
  Department of Computer Science\\
  The University of Chicago\\
  \texttt{\{ady,skr\}@uchicago.edu} \\
}

\begin{document}

\maketitle

\begin{abstract}
Recent work has extensively shown that randomized perturbations of neural networks can improve robustness to adversarial attacks. The literature is, however, lacking a detailed compare-and-contrast of the latest proposals to understand what classes of perturbations work, when they work, and why they work. We contribute a detailed evaluation that elucidates these questions and benchmarks perturbation based defenses consistently. In particular, we show five main results: (1) all input perturbation defenses, whether random or deterministic, are equivalent in their efficacy, (2) attacks transfer between perturbation defenses so the attackers need not know the specific type of defense -- only that it involves perturbations, (3) a tuned sequence of noise layers across a network provides the best empirical robustness, (4) perturbation based defenses offer almost no robustness to adaptive attacks unless these perturbations are observed during training, and (5) adversarial examples in a close neighborhood of original inputs show an elevated sensitivity to perturbations in first and second-order analyses. 
\end{abstract}

\section{Introduction}

The attacks on Convolutional Neural Networks, such as Carlini \& Wagner~\cite{carlini2017adversarial} or PGD~\cite{madry2017towards}, generate
strategically placed modifications to induce a deliberate misprediction.
Recently, there have been many defenses against such attacks that use perturbation during inference time and/or training to improve robustness~\cite{dziugaite2016study, OddsAreOdd, me-net}.
The success of perturbation-based defenses suggests that adversarial examples are not robust themselves, where a small amount of noise can dominate the strategically placed perturbations rendering them ineffective.
However, a detailed understanding of this phenomenon is lacking from the research literature including: (1) what types of perturbations work and what is their underlying mechanism, (2) is this property specific to certain attacks and architectures, (3) whether such defenses are effective and how to make them effective, (4) what the magnitude of perturbations should be, and (5) where to place them in a network.

We first analyze the inference-time input perturbations, where the input to the network is manipulated prior to inference.
The main trade-off is a selection of the perturbation strength to carefully mitigate prediction errors over true examples but maximize recovery of the adversarial examples.
Interestingly enough, we find that this trade-off is consistent across very different families of perturbations, where the relationship between channel distortion (effective perturbation of the input) and robustness is very similar. One might think that a frequency-based or JPEG based defense should be more tuned to a natural image classification setting---but we do not find that this is the case.
The particular distribution of noise added to the inference process is not as important as its magnitude.

The unification of input perturbation defenses gives us an insight into how an attacker might avoid them even if they did not know the particular defense. Our experiments suggest that all the input perturbation defenses are vulnerable to the same types of attack strategy where the attacker simply finds an adversarial example further away from the original image. This optimization procedure can be tuned to find the smallest distance from the original image that closes a ``recovery window'' (demonstrated in our experiments). We can further optimize this distance with a generic attacker that assumes a particularly strong perturbation, based on the additive Laplace noise. Adaptive attacks designed on this channel are often successful against other defenses. This result implies that input perturbation defenses are simply not effective and attackers can easily circumvent them without much knowledge about the particular defense. 

Our experiments compare different placements of noise layers across the network.
In general, we find that the most effective current defenses leverage a strategy used in RobustNet~\cite{RSE}, which adds noise throughout the network.
On the other hand, we show that simply adding noise in a single layer, as in~\cite{DifferentialPrivacyDefense}, is ineffective in practice.
We also find that training the network with this noise is crucial for high accuracy, and there is a threshold of perturbation after which a network must be trained with the noise to achieve any reasonable performance (approximately the standard deviation of a perturbed tensor). Our empirical findings indicate that the injection of noise into internal layers is an important and effective defense that can be combined with adversarial training~\cite{madry2017towards} to further improve robustness.

We believe that our analyses are valuable to the community as they highlight important properties of perturbation-based defenses and the principles of constructing new strong defenses.

\section{Related Work}
Much of the community's current understanding of adversarial sensitivity in neural networks is based on the seminal work by Szegedy et al.~\cite{SzegedyIntriguingProperties}.
Multiple contemporaneous works also studied different aspects of this problem, postulating linearity and over-parametrization as possible explanations~\cite{biggio2013evasion,goodfellow2014explaining}.
Since the beginning of this line of work, the connection between compression and adversarial robustness has been recognized. Researchers noticed this property a few years ago with a number of inference-time ``input perturbation'' defenses, for example, feature squeezing~\cite{xu2017feature}, JPEG compression~\cite{dziugaite2016study}, randomized smoothing~\cite{cohen2019certified}, and types of structured perturbations~\cite{jafarnia2018ppd, zhang2019defending, 2017GuoInputTransformations}.
Other main defense strategies include: 
the idea of defensive network distillation\footnote{The distillation is a form of compression, however, the defensive distillation does not result in smaller models.}~\cite{papernot2015distillation}, quantizing inputs using thermometer encoding~\cite{buckman2018thermometer}, frequency-based compression harnessed by~\cite{Aydemir18,Bafna2018,Das17, Das18,dziedzic2019blt, 2017GuoInputTransformations,Liu19}. 

Another highly related line of research leverages randomization for adversarial robustness: pixel deflection~\cite{Prakash18}, random resizing and padding of the input image~\cite{xie2017mitigating}, total variance minimization~\cite{2017GuoInputTransformations}, dropout randomization~\cite{feinman2017detecting}, random pixel elimination followed by matrix estimation~\cite{me-net}.
More effective perturbation-based defenses are possible by injecting noise to the internal layers~\cite{DifferentialPrivacyDefense, RSE}, model parameters~\cite{2019ParametricNoiseInjection}, or via a separately trained auto-encoder added in front of a network~\cite{DifferentialPrivacyDefense}.
There are also defenses based on random perturbations that give proven guarantees of robustness~\cite{cohen2019certified, DifferentialPrivacyDefense,zhang2019defending}. 
More recent work focuses on a combination of randomized smoothing with adversarial training and achieves state of the art in terms of the provable robustness~\cite{smoothAndAdvTrain2019}. 

Despite all of this research and several theoretical results, the empirical success of these approaches has not completely been established. Many aforementioned defenses were later broken~\cite{athalye2018obfuscated,carlini2017adversarial, tramer2020adaptive}, and many recent defenses have not been compared against each other in a standard-setting.
This motivates our analysis.

\section{Theoretical Background}
\subsection{Notation and Metrics}
We consider convolutional neural networks that take $w \times h$ (width times height) RGB digital images as input, giving an example space of $\mathcal{X} \in (255)^{w \times h \times 3}$, where $(z)$ denotes the integer numbers from $0$ to $z$. We consider a discrete label space of $k$ classes represented as a confidence value $\mathcal{Y} \in [0,1]^k$. Neural networks are parametrized functions (by a weight vector $\theta$) between the example and label spaces $f(x;\theta): \mathcal{X} \mapsto \mathcal{Y}$. An \emph{adversarial input} $x_{adv}$ is a perturbation of a correctly predicted example $x$ that is incorrectly predicted by $f$: $f(x) \ne f(x_{adv})$ The \emph{distortion} is the $\ell_2$ error between the original example and the adversarial one: $\delta_{adv} = \|x - x_{adv}\|_2$. We focus on white-box attacks where the adversary has access to a parametric description of the model he/she is attacking. White-box adversarial examples can be synthesized in three main settings. In the \emph{adaptive} setting, the attacker knows the model and what defense is used.
In the \emph{non-adaptive} setting, the attacker does not know about the defense technique and assumes no defense is used.
Finally, our experiments consider a novel \emph{partially adaptive setting}, where the attacker knows that a perturbation-based defense is being used but does not know precisely which one. 

\subsection{A Unified View on Perturbations}
 While there is extensive work on using randomization or compression as a defense, we find that all of the approaches essentially follow the same format. They approximate a trained network $f(\cdot)$ with a less precise version $\hat{f}(\cdot)$, such that an adversarial example reverts back to the original class: $f(x) = \hat{f}(x_{adv})$. Intuitively, a lossy version of $f$ introduces noise into a prediction which dominates the strategic perturbations found by an adversarial attack procedure. It turns out that we can characterize a number of popular defense methodologies with this basic framework. How and where to inject noise is the core question. We model input perturbations as a general \textit{distorted communication channel} to obtain a unification of these methods. The input perturbations first transform $x$ through a noisy channel $C(x) = C[x' \mid x]$, and then evaluate the neural network $\hat{f}(x) = f(x')$, where $x' \sim C(x)$. One could also perturb the network itself by adding $\epsilon$ noise to the parameters: $\hat{f}(x) = f(x; \hat{\theta})$, where $\hat{\theta} = \theta + \epsilon$ or by injecting noise to any or all of the intermediate layers of a network $f(x) = g(h(x))$, where $\hat{f}(x) = g(h(x + \epsilon_1) + \epsilon_2)$ (Supplement~\ref{sec:model}-\ref{sec:sup-stochastic-perturbations}).

\subsection{Perturbation Analysis}
\label{subsection:perturbationAnalysis}
Part of the goal of this experimental study is to understand the mechanism that allows such defenses to work in the first place.
We start with the hypothesis that synthesized adversarial examples have \textbf{\emph{unstable} predictions--meaning that small perturbations to the input space can change confidence values drastically}---and measure quantities that will allow us to quantify this instability. Let $f(x)$ be a function that maps an image to a single class confidence value (i.e., a scalar output).
We want to understand how $f(x)$ changes if $x$ is perturbed by $\epsilon$.
We can apply a Taylor expansion of $f$ around the given example $x$:
\[
f(x+\epsilon) \approx f(x) + \epsilon^T \nabla_{x}f(x) + \frac{1}{2} \epsilon^T \nabla^2_{x}f(x) \epsilon + ...
\]
where $\nabla_{x}f(x)$ denotes the gradient of the function $f$ with respect to $x$ and $\nabla^2_{x}f(x)$ denotes the Hessian of the function $f$ with respect to $x$. The magnitude of the change in confidence is governed by the Taylor series terms in factorially decreasing importance.
$\delta_{c} = \|\epsilon\|_2$ is distortion measure. The expression is bounded in terms of the \emph{operator} norm, or the maximal change in norm that could be induced, of each of the terms
(see details in Supplement~\ref{appendix:perturbation}):
\[
\epsilon^T \nabla_{x}f(x) + \frac{1}{2} \epsilon^T \nabla^2_{x}f(x)\epsilon + ...
 \leq \delta_{c} ~ M_1(x) + \frac{1}{2} \delta_{c}^2 ~ M_2(x) + ...
\]
As $\nabla_{x}f(x)$ is a vector, this is simply the familiar $\ell_2$ norm, and for the second-order term this is the maximal eigenvalue:
\[
M_1(x) = \|\nabla_{x}f(x)\|_2~~~M_2(x) = \lambda_{max}(\nabla^2_{x}f(x))
\]
When $M_1$ and $M_2$ are larger this means there is a greater propensity to change the prediction for small perturbations.
We will show experimentally that for certain types of attacks the $M_1$ and $M_2$ values around adversarial examples exhibit signs of instability compared to those around natural examples--suggesting a mathematical mechanism of why recovery is possible.

\section{Empirical Results}

\subsection{Experimental Setup}
\label{sec:experimental-setup}
We run our experiments using 
ResNet-18, ResNet-20, or VGG-16 on CIFAR-10, ResNet-20 on SVHN, and ResNet-50 on ImageNet using P-100 GPUs (16GB memory). We explore a number of different attacks that are implemented in the foolbox library~\cite{foolbox}.
In each experiment, we measure the test accuracy (\%), the confidence of predictions, and distances between the original images and either their adversarial counterparts or the recovered images after applying one of the defenses. We present our results for \emph{non-targeted attacks}; if the adversary is successful it induces any misclassification. 



For the adaptive case, we extend attacks to reduce the effects of gradient obfuscation. We approximate the gradients for the backward pass on the compression layers (usually as an identity function), similarly to~\cite{athalye2018obfuscated, he2017adversarial}. For the RobustNet~\cite{RSE} and PNI~\cite{2019ParametricNoiseInjection}, the C\&W attack is aware of the randomization procedure and we use its adaptive version presented in~\cite{RSE}. We also use the PGD attack from~\cite{2019ParametricNoiseInjection} and add EOT~\cite{athalye2018obfuscated}. More details on the setup can be found in Supplement, Section~\ref{sec:setupDetails}.

\begin{SCfigure}[][h]
\includegraphics[width=0.7\columnwidth]{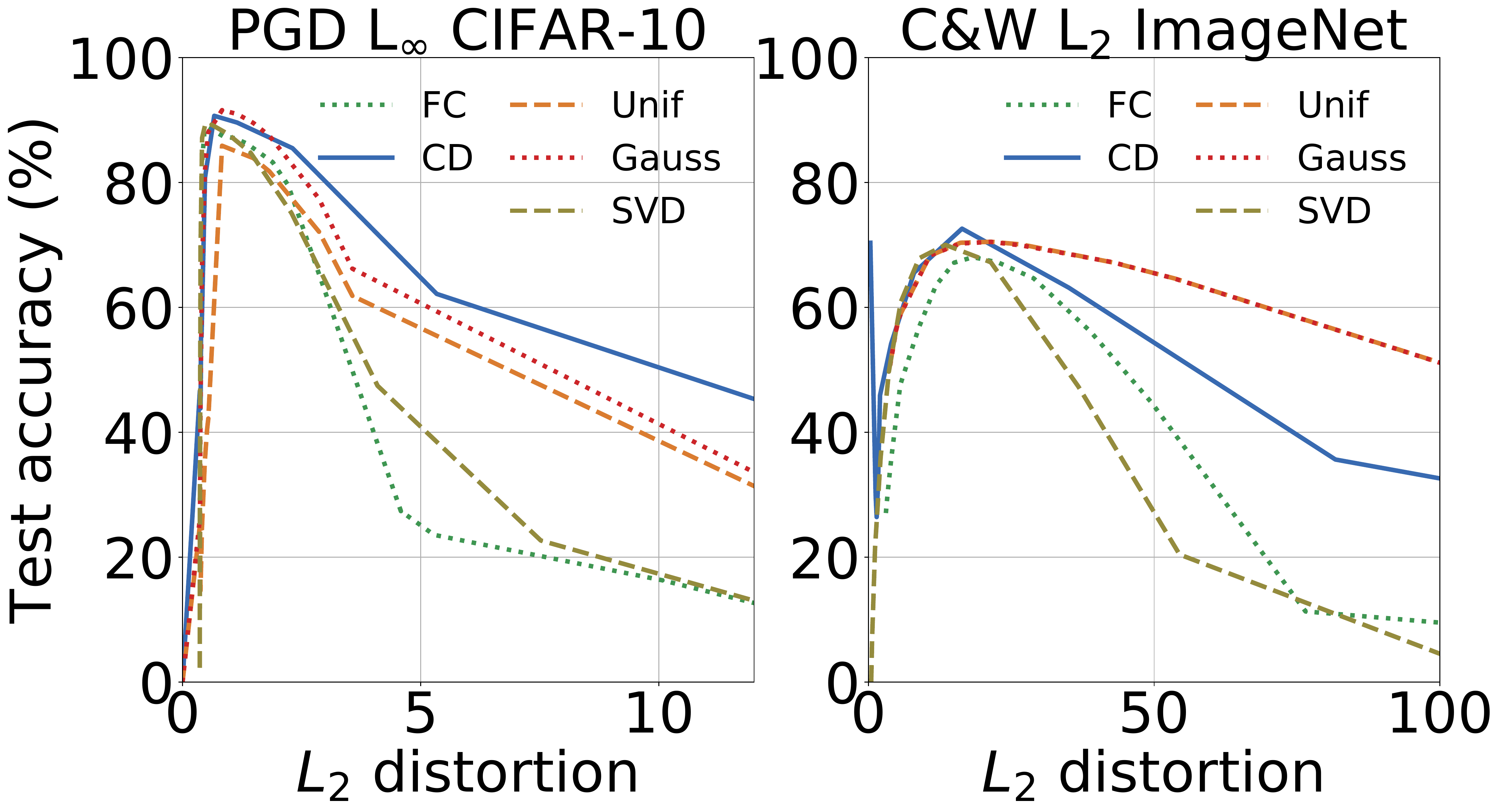}
\caption{\label{fig:full-distortion-vs-accuracy} 
\textbf{\emph{Input perturbations provide similar gains in robustness.}} 
Distortion is due to defense.
Test accuracy on clean data is 93.56\% and 76.13\% for CIFAR-10 (ResNet-18) and ImageNet (ResNet-50), respectively. Defenses: FC - Frequency-based Compression, CD is Color-Depth reduction (feature squeezing), Uniform \& Gaussian noise, SVD compression.
}
\end{SCfigure}


\subsection{Input Perturbation Defenses Lead to Similar Gains in Robustness}
\label{sec:similar-channels}
We first study the simplest class of perturbation defenses that perturb only inputs. In this experiment, we consider the non-adaptive setting where the attacker does not know about the defense. The attack is untargeted, any misclassification is considered a success. An interesting way to compare very different classes of input perturbation defenses is to look at the ``input distortion'' ($L_2$ distance between the input and its perturbation, not to be confused with the distortion of the attack). We compare this metric against the test accuracy of a defended model, which indicates the ability of the perturbation to recover the original label.

For each image from the CIFAR-10 test and ImageNet dev sets, we generate an adversarial attack using PGD (40 iterations with a random start) and C\&W (100 iterations). The attacks are tuned to minimize the distance between adversarial inputs and original examples.
We present the results in Figure~\ref{fig:full-distortion-vs-accuracy}. 
$L_2$ distortion of 0.0 indicates no defense and the attacks reduce the test accuracy to 0\%. The defenses are applied only during inference. Each of the tested perturbation techniques can recover a substantial portion of correct labels from the adversarial examples. They recover the accuracy to about 85\% for \mbox{CIFAR-10} and 70\% for ImageNet. The curves are very similar in terms of where they achieve their peak robustness---the particular distribution of perturbations is not as relevant as their magnitude. The ``optimally-tuned'' defense essentially finds the same distortion parameter across very different stochastic and deterministic techniques.
\subsection{How to Attack Input Perturbation Defenses Non-adaptively?}
\label{sec:recoveryWindow}
Input perturbation defenses might seem a simple and effective trick to make a model adversarially robust, however, their similarity is a major pitfall. Even if the attacker does not know which particular defense is being used he/she can still attack the model with a ``less-precise'' attack---one that creates a larger distortion but higher confidence adversarial image. The experiment in Section~\ref{sec:similar-channels} shows that the defense philosophy is to generate a perturbation big enough to dominate the adversarial perturbations but small enough to generate valid predictions. 
This creates an easy to recognize attack vector, where the attacker simply makes the adversarial perturbations large enough that the defender significantly hurts the accuracy of the model when trying to dominate the adversarially placed strategic perturbations. Such attacks remain imperceptible (Supplement~\ref{subsection:NeighborhoodAdversarialExamples}).

\subsection{Partially Adaptive Setting}
\label{sec:partially-adaptive-setting}

\begin{SCtable}
\centering
\caption{\textbf{\emph{Transferability of adversarial images}} created using partially adaptive attack (\textit{A}) and tested against defense (\textit{D}). Each cell represents a recovered test accuracy (\%). The clean test accuracy is 93.56\% for CIFAR-10 dataset trained on ResNet-18 architecture. 
}
\label{table:attack-transfer}
\begin{tabular}{ccccccc}
\toprule
\backslashbox{\textit{A}}{\textit{D}} & FC & CD & SVD & Gauss & Uniform & Laplace \\
\midrule
\textit{Empty} & 93.32 & 93.01 & 93.12 & 92.53 & 91.6 & 91.35 \\
\hline
FC & \textbf{0.20} & 80.75 & 83.05 & 81.15 & 79.65 & 78.70 \\
CD & 3.85 & \textbf{0.70} & 43.60 & 47.30 & 60.45 & 62.35 \\
SVD & 1.99 & 47.96 & \textbf{0.77} & 46.52 & 62.87 & 65.75 \\
\hline
Gauss &    4.45 & 48.70 & 44.80 & 51.50 & 61.75& 60.15 \\
Uniform & 3.45 & 30.30 & 30.60 & 30.15 & 48.05 & 51.55 \\
Laplace & 3.05 & 23.35 & 24.60 & \textbf{23.80} & \textbf{39.15} & \textbf{46.70} \\
\toprule
\end{tabular}


\end{SCtable}

The perturbation-based defenses fail if the attacker makes the adversarial distortion large enough. Can an attacker still break such defenses with a low-distortion attack? One approach is to consider the adaptive setting. If the attacker has full knowledge of the defense, it is possible to construct an (adaptive) attack that is impervious to the defense. It has been known that many perturbation-based defenses are easily broken in the adaptive setting.

We go one step further. Since our experiments show that input transformations are so similar in their mechanisms, we find that an attacker \emph{does not need to be fully adaptive}. To the best of our knowledge, such a problem setting has not been studied in the previous literature. The attacker simply assumes a particular strong perturbation-based defense and that same adversarial input often transfers to other defenses. 
Our attacker assumes that the defender is using a Laplace noise perturbation to defend the model, and generates an attack.
We restrict the attacker to a single adaptive step (for details see Section~\ref{white-box-adaptive-attack-sup} in Supplement). Even in this weak adaptive setting, the deterministic channels are fully broken, and the randomized channels are mostly broken (with a maximum accuracy of about 23.8\%). We can strengthen this attack by giving to an adversary a larger ``compute budget'', for example, with an unlimited number of adaptive steps~(Figure~\ref{fig:manyIters} in Supplement) and thereby driving the accuracy to 0.


Laplace attacked images transfer the best to other defenses (Table~\ref{table:attack-transfer}). They decrease the accuracy of the defense models by at least 44.3\% (for the Laplace-based defense itself). 
FC attacked images do not transfer well to other defenses and the maximum drop in accuracy of the model protected by other defenses is 12.26\%.
Most adversarial images (against a given defense) transfer very well to the FC defense. An adversarial image against any defense (e.g. CD, SVD, Gauss, Uniform, or Laplace) is also adversarial against the FC defense. 
The adversarial images generated against the Uniform defense show better transfer to other defenses in comparison to the adversarial images generated against the Gaussian defense. This is because the higher noise level is applied in the Uniform defense. We observe analogous trends for the ImageNet dataset (Supplement~\ref{Sec:SupTransferability}).

\subsection{Robustness to Adaptive Attacks}

\begin{figure}[htb!]
    \centering
    \includegraphics[width=1.0\columnwidth]{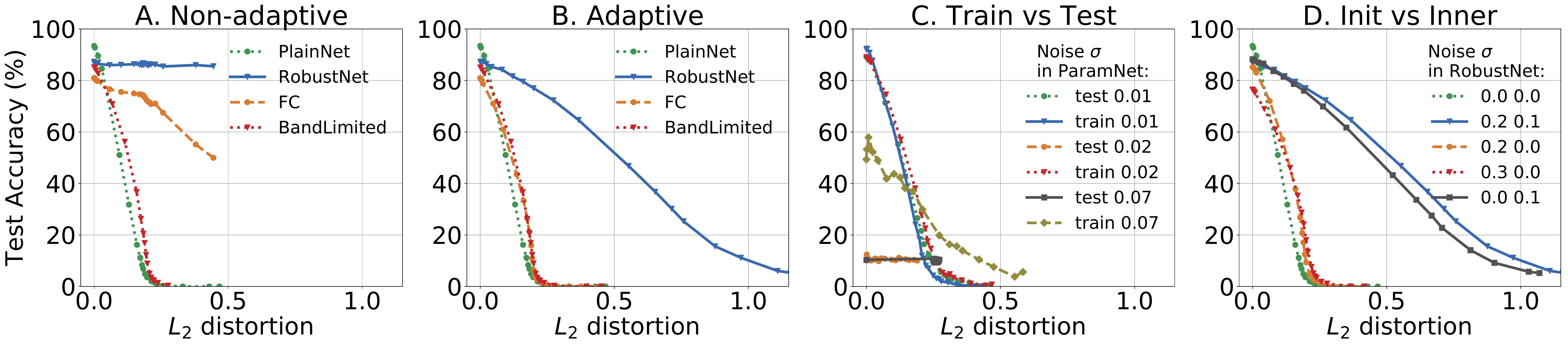}
    \caption{
    \label{fig:adaptive-param-robustnet}
    \emph{Comparison between attacks.}
    \emph{\textbf{A. Non-adaptive attack.}} Adversarial images  generated on PlainNet and tested against RobustNet, FC, and BandLimited defenses. \emph{\textbf{B. Adaptive.}} Generate the attack with knowledge about the defenses. 
    \emph{\textbf{C. Train vs Test.}} The robustness of ParamNet with noise added either during both train and test phases or only during test. \emph{\textbf{D. Init vs Inner.}} Higher robustness when noise layers placed before every convolutional layer (0.2 0.1) than only before the first layer (0.2 0.0 and 0.3 0.0) or only in internal layers (0.0 0.1). We use C\&W L2 attack and VGG-16 trained on CIFAR-10. 
   }
\end{figure}


The previous set of experiments shows that simply adding post-hoc perturbations to a model is not an effective defense. A crucial way to boost the robustness of perturbation based defenses is to \emph{train the model} observing training data perturbed by the future defense. This approach seems similar to adversarial training but turns out to be much more computationally efficient as the perturbations are usually easy to calculate. While it has been previously noted that adversarial examples can fool multiple models~\cite{tramer2017space}, we find that perturbation defenses significantly affect transferability.

We test FC -- the frequency-based channel with a 50\% compression rate, which is one of the better performing input perturbations in this setting.
RobustNet~\cite{RSE} adds a random noise layer in front of each convolutional layer (we set the noise in the initial layer (init) to 0.2, and in all the internal layers (inner) to 0.1 as in~\cite{RSE}). The BandLimited model~\cite{dziedzic2019blt} modifies each convolutional layer by using FFT-based convolutions with compression (we set the compression rate to 80\%). 

For the non-adaptive attack, adversarial images generated on PlainNet do not transfer to RobustNet (see Figure~\ref{fig:adaptive-param-robustnet} A). On the other hand, the BandLimited model, which is robust to Gaussian noise, is not robust to gradient-based attacks. We observe that gradients estimated for the BandLimited convolution operations are approximate enough and thus useful to generate the first-order attacks. 
The FC defense provides even up to 50\% robust accuracy in the nonadaptive setting (for $L_2$ distortion of 0.4), however, once we compute its gradients (similarly to the BandLimited layers) in the adaptive setting, the accuracy of the defense drops to 0\% (Figure~\ref{fig:adaptive-param-robustnet} B). 

In general, many defenses rely on gradient obfuscation. The latest work~\cite{tramerNips2019} indicates that adversarial training suffers from gradient masking as well. Thus, we extend the adaptive case and test RobustNet against PGD + EOT (Expectation Over Transformation)~\cite{athalye2018obfuscated}. Attacking RobustNet with 10 iterations of EOT decreases its accuracy by about 10\% in comparison to the standard PGD attack. However, there are diminishing returns for more EOT iterations (Figure~\ref{fig:pgd-eot} in Supplement). Thus, EOT helps partially but does not allow us to find fully useful gradients to defeat the defense. This result implies that RobustNet obfuscates the gradients by randomizing them. 
Another way to identify gradient obfuscation is to check if a defense performs better against white-box than black-box attacks. It occurs that, for example, the Boundary black-box attack~\cite{brendel2017decision} is unsuccessful since it executes a random walk along (in this case) a \textit{random decision boundary} (Supplement~\ref{sec:boundary-attack}).

\subsection{Noise Injection during Train vs Test Time}
\label{sec:noise-injection}

RobustNet randomly perturbs inputs and feature maps. An alternative approach is to randomly perturb parameters and we call such network a ParamNet. 
We investigate how much noise can be added to ParamNet so that the clean accuracy does not decrease significantly and the robustness of the network is improved (Figure~\ref{fig:adaptive-param-robustnet} C). For a small amount of noise ($\sigma=0.01$), the drop in clean accuracy is negligible, however, there is no gain in robustness. Interestingly enough, a slight increase in the amount of injected noise (to $\sigma=0.02$) during the test phase turns the network into a random classifier. This problem arises when the values of the parameters are dominated by random noise. To restore clean accuracy, it is necessary to train the network with noise.
We can improve robustness by further increasing the amount of noise ($\sigma=0.07$) but at the cost of lower accuracy on clean data.


Figure~\ref{fig:adaptive-param-robustnet} D demonstrates where and how to add noise to RobustNet. 
Adding noise to the first layer is essentially the same as input perturbation. We could also add noise only to the internal layers~\cite{DifferentialPrivacyDefense}, however, to achieve a high level of robustness, we have to add noise to all the layers. 
We also test different types of noise: Gauss, Uniform, and Laplace injected into RobustNet. Following the general rules established in the preceding Section~\ref{sec:similar-channels}, the robustness of the network is independent of the type of injected noise (Figure~\ref{fig:robustnet-gauss-uniform-laplace} in Supplement). 

RobustNet can be extended to an ensemble during test time by performing several forward propagations, each time with different prediction scores due to the noise layers. However, we argue that ensemble learning is not an appropriate way of increasing robustness to adversarial attacks. In general, the ensemble method performs better than a single model if each of the ensemble models is a good predictor. On the other hand, when a single predictor does not perform well then the ensemble of the weak predictors magnifies the bad effect and makes the ensemble model worse than a single predictor. For weak attacks, RobustNet performs well and its good performance is amplified by the ensemble. For strong attacks, the RobustNet predictor performs poorly and ensemble magnifies the errors.
Additionally, from the system perspective, the ensemble for RobustNet decelerates the inference process.

It occurs that RobustNet is significantly more robust than ParamNet (Figures~\ref{fig:adaptive-param-robustnet} C and D). The reason is two-fold. First, the number of parameters is an order of magnitude lower than the number of elements in the input and intermediate feature maps~\cite{2016vDNNnvidia}. Second, the standard deviations of tensors for features are an order of magnitude higher than for parameters. Thus, a more effective amount of variation can be added to features than to parameters.

\begin{figure*}[t]
    \centering
    \includegraphics[width=1.0\linewidth]{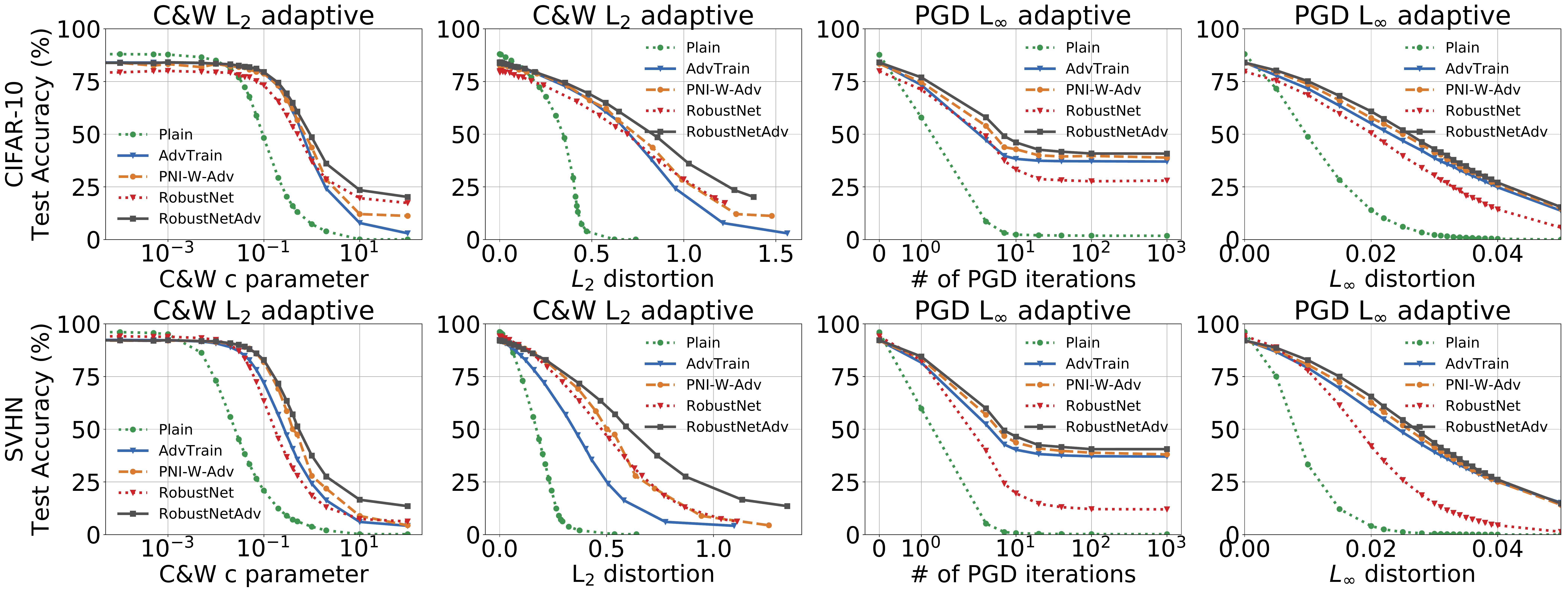}
    \caption{\label{fig:adaptive-pgd-cw} White-box adaptive attacks C\&W and PGD used to generate adversarial examples and tested against the standard model (Plain) as well as the following defenses: adversarial training (AdvTrain), parameter noise injection to the weights that uses adversarial training (PNI-W-Adv), feature noise injection (RobustNet), and RobustNet combined with adversarial training (RobustNetAdv). We train ResNet-20 on CIFAR-10 and SVHN datasets.}
\end{figure*}


\subsection{Noise Injection and Adversarial Training}
Adversarial training (AdvTrain)~\cite{madry2017towards} is one of the most successful practical defenses. RobustNet is another strong defense that injects noise into inputs and intermediate feature maps~\cite{RSE}. PNI-W-Adv~\cite{2019ParametricNoiseInjection} combines noise injection into network parameters with adversarial training and can give higher robustness than the pure AdvTrain defense. Previous Section~\ref{sec:noise-injection} shows that injecting noise into network layers performs better than noise injection into parameters. Thus, we propose to combine RobustNet with AdvTrain (denoted as RobustNetAdv) and compare it to AdvTrain, PNI-W-Adv, and RobustNet defense methods.

\textbf{Attack Setup.} We run the C\&W attack with 200 iterations for the gradient descent and vary the $c$ parameter from 0 to 100. For all compared defenses, adversarial training uses PGD with 7 iterations, similarly to~\cite{2019ParametricNoiseInjection, madry2017towards}, the perturbation scale of 8/255 (0.031), and step size 2.55/255 (0.01). We only vary either the number of iterations for the attack or perturbation scale $\epsilon$ (which influences the $L_\infty$ distortion of adversarial examples). Besides using 40 and 100 total attack iterations, we also increase the iterations to 1000 to further strengthen the adversary, similarly to~\cite{me-net}. The current state of the art attack is white-box and adaptive. RobustNet and PNI-W-Adv employ the randomization techniques so we use a modified version of attacks with EOT (Expectation Over Transformation)~\cite{athalye2018obfuscated}. 

\textbf{Noise injection requires manual fine-tuning.}
We tune noise layers in RobustNet. The standard deviations of inputs to convolutional layers are guiding values for the injected noise level. To achieve good performance, we adjust the noise separately for at least the initial and internal layers.
PNI-W-Adv goes a step further and injects noise scaled by a \textit{trained factor}. The initial noise magnitude is based on the standard deviation of network parameters. However, it requires us to adjust weights in ensemble loss, which consists of two terms for clean data loss and adversarial data loss.
Training RobustNet in a similar manner to PNI-W-Adv usually lowers the injected noise scale too much and makes the network much less robust.

\textbf{Defense Setup}. For RobustNet, we set the standard deviation $\sigma$ of the injected noise to 0.2 (and 0.08) in the input layer and 0.1 (and 0.07) in all the remaining internal layers for CIFAR-10 (and SVHN). For RobustNetAdv, we keep the same $\sigma=0.1$ for all the noisy layers for CIFAR-10 and use the same 0.08 for initial and 0.07 for internal layers for SVHN. We train PNI-W-Adv in the same setup as in~\cite{2019ParametricNoiseInjection}, with the equally weighted (by 0.5) sum of losses for clean and adversarial data.

We present the results in Fig.~\ref{fig:adaptive-pgd-cw} as the test accuracy being a function of either strength of an attack or its incurred distortion. The former approach allows us to adjust attack parameters while the latter allows direct comparison between defenses. 

\textbf{A more robust network requires adversarial examples with higher distortion.} For the same $c$ parameter in CW, the attack generates different adversarial examples depending on the attacked network. For example, for parameter $c=0.4$, the distortion required to attack RobustNet is 0.63, whereas to attack AdvTrain, the required distortion is 0.57. 

\textbf{Combining adversarial training with noise injection is beneficial.} AdvTrain and PNI-W-Adv are trained against PGD and for this attack perform better than RobustNet. However, for CW, RobustNet outperforms AdvTrain. Then, RobustNetAdv provides an improvement also in comparison to PNI-W-Adv for CIFAR-10 and SVHN datasets when using CW or PGD attacks. PNI-W-Adv is on par with AdvTrain for CW for most distortion levels and slightly outperforms it for highly distorted examples, similarly to RobustNet. The placement of the noise injections is important and RobustNetAdv outperforms PNI-W-Adv in most cases.

\subsection{Adversarial Examples Are Unstable}
\label{sec:unstable-adv-examples}


The key question is why adversarial examples are more sensitive to perturbations than natural inputs when there is evidence that from an input perspective they are statistically indistinguishable.
Our experiments suggest that this sensitivity arises from the optimization process that generates adversarial examples.

Based on the operator-norm analysis presented in Section~\ref{subsection:perturbationAnalysis}, we measure $L_2$ norm of the input gradients w.r.t. the original $x_{org}$ and adversarial $x_{adv}$ images for original (correct) $c_{org}$ and adversarial classes $c_{adv}$. 
For natural images, the class with the lowest input gradient is always the class with the highest confidence.
Figure~\ref{fig:attack_conf_cw} (in Supplement) shows that the lower the attack distortion, the more likely a ``gradient anomaly'' for the adversarial example, where the lowest gradient does not correspond to the highest confidence class.
This is not surprising in retrospect---at first-order approximation, an $\epsilon$-sized $L_2$ \textit{untargeted} adversarial attack increases the loss $\mathcal{L}$ at point $x$ by $\epsilon \|\partial_x \mathcal{L}(x, c_{org})\|_2$. Analogously, at first-order approximation, an $\epsilon$-sized $L_2$ \textit{targeted} adversarial attack decreases the loss $\mathcal{L}$ at point $x$ by $\epsilon\|\partial_x \mathcal{L}(x, c_{adv})\|_2$~\cite{2019SimonGabrielFirstOrder}. We can systematically measure the phenomena by adding more Gaussian noise to the original or adversarial images (Figures~\ref{fig:gauss_grads},~\ref{fig:gauss_grads_adv} in Supplement).

We extend the second-order analysis~\cite{HessianBatch} and compute Hessians with respect to inputs instead of parameters. Our experiments show that adversarial examples lead to an order of magnitude higher eigenvalues of the Hessians (on average) than original inputs (Figure~\ref{fig:hessian-cifar10} in Supplement). This suggests that the model predictions for the adversarial inputs are less stable and random perturbations of the adversarial images with some form of noise can easily change the classification outcome. On the other hand, predictions for the original images are more stable and retain their correct labels when perturbed with a small amount of random noise.

\section{Conclusions}
The non-adaptive attacks are not robust since small changes to adversarial inputs often recover the correct labels. This is an obvious corollary to the very existence of adversarial examples that by definition are relatively close to correctly predicted examples in the input space. Random perturbations of inputs can dominate the strategically placed perturbations synthesized by an attack.
The results are consistent across both deterministic and stochastic channels that degrade the fidelity of input examples. From the perspective of the attacker, the recovery window can be closed to make the perturbation based techniques ineffective. Moreover, a strong input perturbation defense can be assumed a priori to achieve high transferability of the attacks. The most effective defenses perturb not only inputs but also the internal layers. The scale of the noise can be set as a parameter. We find that the combination of layer perturbations and adversarial training improves robustness against adaptive CW and PGD attacks in comparison to pure adversarial training or parameter noise injection with adversarial training.  
The training of networks with strong perturbation defenses (high level of added noise) is essential to recover the test accuracy on clean data. Our first and second-order analyses show that the perturbation based defenses are possible since the adversarial examples are not robust themselves. 

\section*{Broader Impact}
The perturbation analyses along with the results in Section~\ref{sec:unstable-adv-examples} can be useful for further theoretical investigation of the source of instability of adversarial examples. This paper should be helpful for practitioners in designing adversarial defenses in the future. On the other hand, the exposed attacks might put a curb on the adaptation of machine learning systems, especially those that could directly interact with human beings. Our method does not leverage any biases in the data.



\bibliography{bibliography}
\bibliographystyle{abbrv}

\appendix
\newpage

\section{Robustness to Black Box Attack}
\label{sec:boundary-attack}
The zero-order methods do not leverage model gradients and are usually used to attack models that we do not have direct access to. They also help to assess if a given model obfuscates gradients. We test our models against a state-of-the-art black-box Boundary attack~\cite{brendel2017decision}, that intuitively is based on a random walk along a decision boundary. 

It is not trivial to find an initial random adversarial example when attacking RobustNet (or PNI-W-Adv). The Boundary attack is initialized by random adversarial examples drawn from a uniform distribution. The uniform noise is applied many times (subject to a hyper-parameter) in different directions. After an initial adversarial example is found, the iterative interpolation is used between the found random adversarial example and its corresponding original image by gradually moving from the original image towards the adversarial example until the interpolated example is misclassified. However, RobustNet applies random perturbations during inference, which becomes a non-deterministic process that can classify differently the same image during consecutive tests. Thus, initially found random adversarial examples might not remain adversarial for the next step of the attack. This hinders the proper initialization of the Boundary black-box attack and causes that only about 20\% of the initially found adversarial examples remain adversarial for the main part of the algorithm. 

\begin{SCfigure}
    \centering
    \includegraphics[width=0.43\columnwidth]{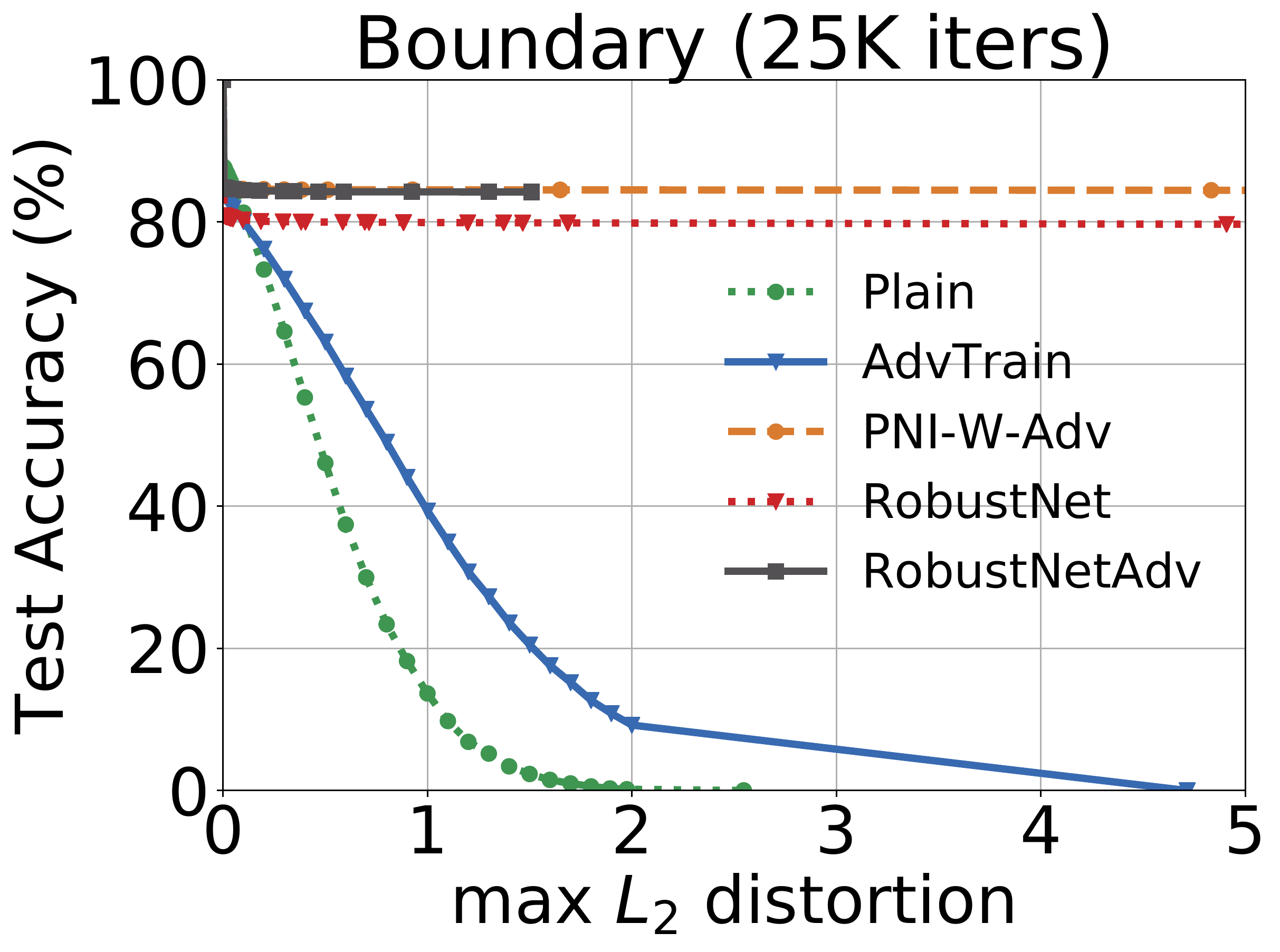}
    \caption{\emph{\textbf{Robustness to black-box attack.}}
    We use the Boundary attack with 25K iterations tested against the standard model (Plain) as well as the following defenses: adversarial training (AdvTrain), parameter noise injection to the weights that uses adversarial training (PNI-W-Adv), feature noise injection (RobustNet), and RobustNet combined with adversarial training (RobustNetAdv). We train ResNet-20 model on CIFAR-10 dataset.
    \label{fig:black-box}}
\end{SCfigure}

To ensure a high rate of initial adversarial examples, we run the initialization until all examples in a given batch are adversarial and apply 25K iterations for the main part of the algorithm, in which random directions are explored and a similar issue to the initialization repeats. This could be partially ameliorated by testing the same random direction many times but it makes the already expensive algorithm a few times more costly. The Boundary attack optimizes for the $L_{2}$ distance between the adversarial and original examples, so we plot the test accuracy (\%) against a wide range of possible $L_{2}$ distortions (from 0.0 to 5.0). We present the results in Figure~\ref{fig:black-box}. 

The parameter noise injection to the weights (PNI-W-Adv) performs similarly to the RobustNetAdv defense. Thus, an appropriate perturbation (either in parameter or feature map space), combined with adversarial training, gives a more robust defense than either of the methods alone. The RobustNet and PNI models are randomized models and their decision boundary is also stochastic. Thus, Boundary attack executes the random walk on the random decision boundary and this approach fails to evade RobustNet.

\section{Details on the experimental setup}
\label{sec:setupDetails}
We use the foolbox library~\cite{foolbox} and borrowed the nomenclature from there. 
In most of our experiments, we use the default foolbox parameters for the attacks. For example, for PGD the initial limit on the perturbation size epsilon is set to 0.3, step size to 0.01, the default number of iterations is 40. For Carlini \& Wagner, we set the maximum number of iterations to 1000, the learning rate to 0.005, the initial value of the constant $c$ is 0.01. Note that for the Carlini \& Wagner attack, we use the $c$ parameter as described in~\cite{carlini2017adversarial} and also code from~\cite{RSE}. 

\section{Model}
\label{sec:model}
Approximating $f(\cdot)$ with a less precise version $\hat{f}(\cdot)$ can counter-intuitively make it more robust
~\cite{dziugaite2016study}:
\[
 f(x) = \hat{f}(x_{adv})
\]
Intuitively, a lossy version of $f$ introduces noise into a prediction that dominates the strategic perturbations found by an adversarial attack procedure.
It turns out that we can characterize many popular defense methodologies with this basic framework.

Let $x$ be an example and $f$ be a trained neural network.
Precise evaluation means running $f(x)$ and observing the predicted label.
Imprecise evaluation involves first transforming $x$ through a deterministic or stochastic noise process $C(x) = C[x' \mid x]$, and then evaluating the neural network
\[
y = f(x')~~~~~x' \sim C(x)
\]
We can think of $C(x)$ as a noisy channel (as in signal processing).
The \emph{distortion} of a $C(x)$ is the expected $\ell_2$ reconstruction error:
\[
\delta_{c} = \mathbf{E}[\|C(x) - x \|_2],
\]
which is a measure of how much information is lost passing the example through a channel.

This paper shows that there is a subtle trade-off between $\delta_{c}$ and $\delta_{adv}$. In particular, we can find $\delta_{c}$ such that $\delta_{c} >> \delta_{adv}$ and $f(x) = f(C(x_{adv}))$. 
We show that compression and randomization based techniques exhibit this property.

\section{Deterministic perturbation techniques}
\label{sec:supplement-compression-techniques}

\subsection{Color-depth compression}
When $C(x)$ is a deterministic channel it can be thought of as a lossy compression technique.
Essentially, we run the following operation on each input example:
\[
x' = \texttt{compress}(x)
\]
One form of compression for CNNs is color-depth compression.
The most common image classification neural network architectures convert the integer-valued inputs into floating-point numbers. We abstract this process with the \texttt{norm} function that for each pixel $n \in (255)$ maps it to a real number $v \in [0,1]$ by normalizing the value and the corresponding \texttt{denorm} function that retrieves the original integer value (where $\lfloor\rceil$ denotes the nearest integer function)~\footnote{More complex normalization schemes exist but for ease of exposition we focus on this simple process.}:
\[
\texttt{norm}(n) := \frac{n}{255} ~~~~~~~~ \texttt{denorm}(v) := \lfloor 255*v \rceil
\]
This process is reversible $v = \texttt{norm}(\texttt{denorm}(v))$, but we can artificially make this process lossy. Consider a parametrized $C(\cdot)$ version of the color-depth compression function:
\[
C(v, b) := \frac{1}{2^b-1} \cdot \lfloor (2^b-1)*v \rceil
\]
By decreasing $b$ by $\Delta b$ we reduce the fidelity of representing $v$ by a factor of $2^{\Delta b}$ (for the $b$ bits of precision). 

\subsection{FFT-based compression}
We apply compression in the frequency domain to reduce the precision of the input images.
Let $x$ be an input image, which has corresponding Fourier  representation that re-indexes each tensor in the frequency domain:
\[
F[\omega] = F(x[\mathbf{n}]) 
\] 
This Fourier representation can be efficiently computed with an FFT.
The mapping is invertible $x = F^{-1}(F(x))$. 
Let $M_f[\omega]$ be a discrete indicator function defined as follows:
\[
M_f[\omega] = \begin{cases}
1, \omega \le f\\
0,\omega > f
\end{cases}
\]
$M_f[\omega]$ is a mask that limits the $F[\omega]$ to a certain \emph{band} of frequencies.
$f$ represents \emph{how much of the frequency domain} is considered.
The \emph{band-limited} spectrum is defined as, $F[\omega] \cdot M_f[\omega]$,
and the band-limited filtering is defined as:
\[  x' = F^{-1}(F[\omega] \cdot M_f[\omega]) 
\]

\subsection{SVD-based compression}
Analogously to the FFT-based method, we decompose an image with SVD transformation and reconstruct its compressed version with dominant singular values. The bases used in SVD are adaptive and determined by an image, as opposed to the pre-selected basis used in FFT. This can result in a higher quality for the same compression rate in the case of SVD, however, it is more computationally intensive than FFT-based compression.

\section{Stochastic Perturbations}
\label{sec:sup-stochastic-perturbations}
The channel model is particularly interesting when $C(x)$ is stochastic.
Randomization has also been noted to play a big role in strong defenses in prior work~\cite{cohen2019certified, RSE, DifferentialPrivacyDefense,madry2017towards, zhang2019defending}.
For example, we could add independent random noise to each input pixel or an internal feature map (intermediate layer):
\[
x' = x + \epsilon
\]
We consider Gaussian $\epsilon \sim N(\mu=0,\sigma)$ (with zero mean $\mu$ and $\sigma$ standard deviation), additive Uniform noise $\epsilon \sim U(-B,B)$ (with bound $B$), and Laplace noise $L(\mu=0, b)$ (with zero mean, and scale $b$), which are added independently to each pixel or an element in a feature map. One of the advantages of randomization is that an adversary cannot anticipate how the particular channel $C$ will transform an input before prediction. 


\section{Perturbation analysis: addendum}
\label{appendix:perturbation}
\[
\epsilon^T \nabla_{x}f(x) + \frac{1}{2} \epsilon^T \nabla^2_{x}f(x)\epsilon + ...
 \leq \delta_{c} ~ M_1(x) + \frac{1}{2} \delta_{c}^2 ~ M_2(x) + ...
\]
\[
M_1(x) = \|\nabla_{x}f(x)\|_2~~~M_2(x) = \lambda_{max}(\nabla^2_{x}f(x))
\]

\[
(1) ~~~ \epsilon^T \nabla_{x}f(x) \leq \delta_{c} ~ M_1(x) 
\]
\[
\text{From the Cauchy-Schwarz inequality:\\}
\]
\[
\epsilon^T\nabla_x f(x) \leq ||\epsilon||_2 ~ M_1(x) 
\]
\[
||\epsilon||_2 ~ M_1(x) = \delta_{adv} ~ M_1(x) \le \delta_{c} ~ M_1(x) ~~(\text{ since } \delta_{adv} << \delta_{c})
\]

\[
(2) ~~~ \nabla^2_{x}f(x)\epsilon \le \delta_{c}^2 ~ M_2(x)
\]
\[
\text{From the definition of maximum eigenvalue}: 
\]
\[
\lambda_{max} \ge \frac{\epsilon^T \nabla^2_{x}f(x)\epsilon}{\epsilon^T\epsilon}
\]
\[
\epsilon^T \nabla^2_{x}f(x)\epsilon \le ||\epsilon||_2^2\lambda_{max} = \delta_{adv}^2\lambda_{max} \le \delta_{c}^2\lambda_{max}
\]

\section{Accuracy of Perturbation Defenses on Clean Data}
\label{sec:acc-clean}
One pitfall of the input perturbation defense is that it introduces errors whether or not there are any adversarial examples. The errors act as an upper-bound for the best possible test accuracy we can get under adversarial perturbations. 

Table \ref{max-acc-label} shows the results for all test images from CIFAR-10 on the ResNet-18 architecture, for three of the imprecise channels, and for different noise settings.
\begin{table}[t]
\centering
\caption{On CIFAR-10 with ResNet-18, we measure the max test accuracy without any adversarial perturbation. This signifies the amount of accuracy we sacrifice with respect to the baseline clean test accuracy 93.56\% of the model (without any perturbations of the images).}
\label{max-acc-label}
\begin{tabular}{cc}
\toprule
FC (\%)  & Acc. (\%) \\
\midrule
1  & 93.5     \\
10  & 93.42  \\
50  & 91.6    \\
75  & 79.53           \\
\toprule
\end{tabular}
\hspace{1em}
\begin{tabular}{cc}
\toprule
CD (bits)  & Acc. (\%) \\
\midrule
8  & 93.4     \\
6  & 93.3  \\
4  & 91.9    \\
2  & 87.4           \\
\toprule
\end{tabular}
\hspace{1em}
\begin{tabular}{cc}
\toprule
Uniform ($\epsilon$) & Acc. (\%) \\
\midrule
$0.009$  & 93.52 \\
$0.03$ & 92.59 \\
$0.07$  & 85.2 \\
$0.1$  & 70.67 \\
\toprule
\end{tabular}
\end{table}

\begin{figure}[t]
    \centering
    \includegraphics[width=\columnwidth]{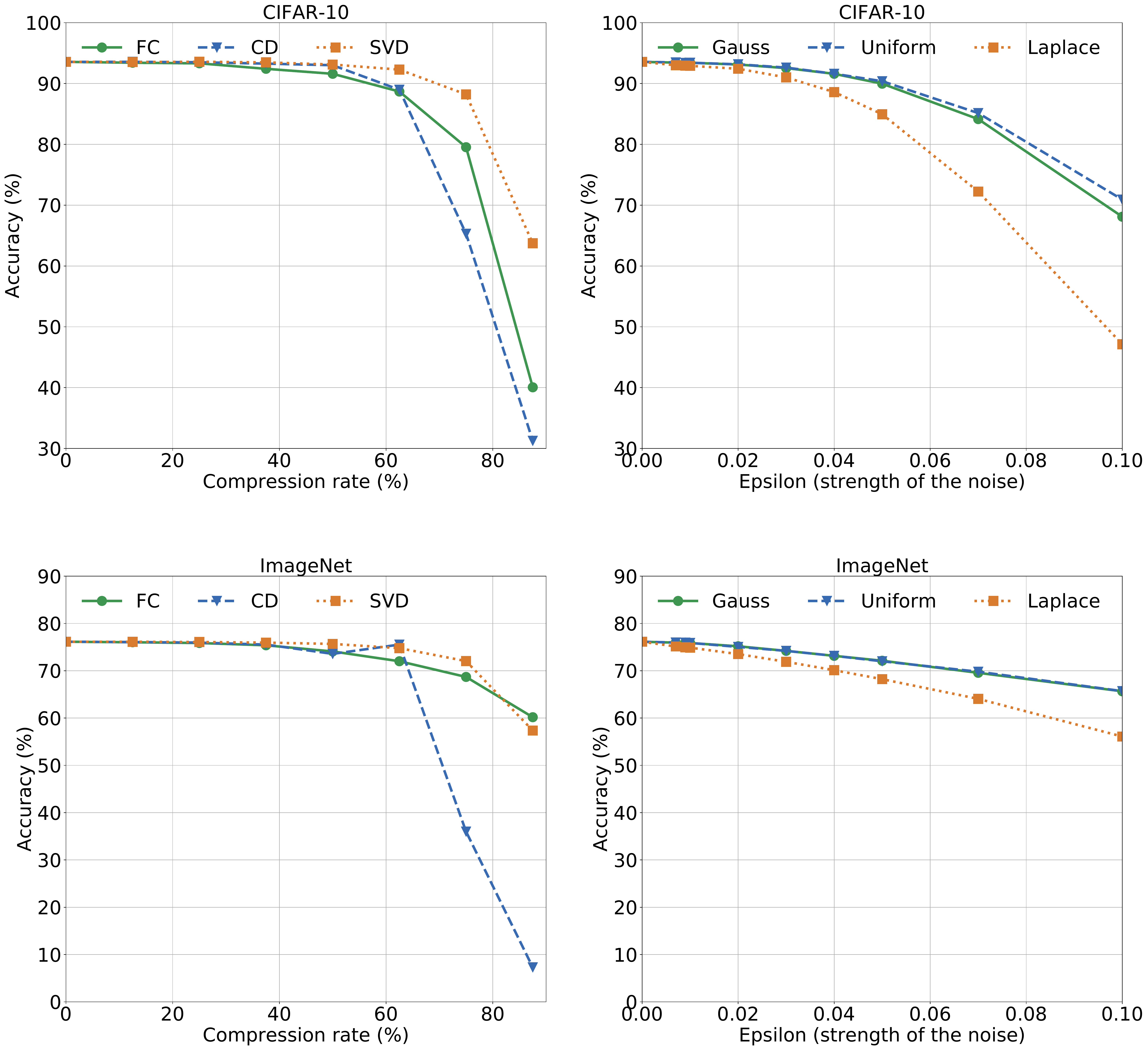}
     \caption{\label{fig:max-test-accuracy-full} The test accuracy after passing clean images through six different input perturbation defenses, where the added noise is controlled by the compression rate and epsilon parameters. We use full CIFAR-10 test set for ResNet-18, and full ImageNet validation set for ResNet-50.}
\end{figure}

We present the results in Figure~\ref{fig:max-test-accuracy-full} for six different noisy channels; three of them are compression based: FC, CD, SVD, and other three add different types of noise: Gauss, Uniform, and Laplace. For each of the compression based approaches, we increase the compression rate systematically from 0 to about 90\% (in case of the CD, the compression rate is computed based on how many bits are used per value). For the noise-based approaches, we increase the strength of the noise by controlling the epsilon parameter $\epsilon$ (in case of the Gaussian noise, it corresponds to the standard deviation parameter $\sigma$). 

The test accuracy of the models can be increased by training with compression, e.g., by using FFT based convolutions with 50\% compression in the frequency domain increases the accuracy to 92.32\%.

\section{Distributions of input perturbations}

\begin{figure}[t]
    \centering
    \includegraphics[width=1.0\columnwidth]{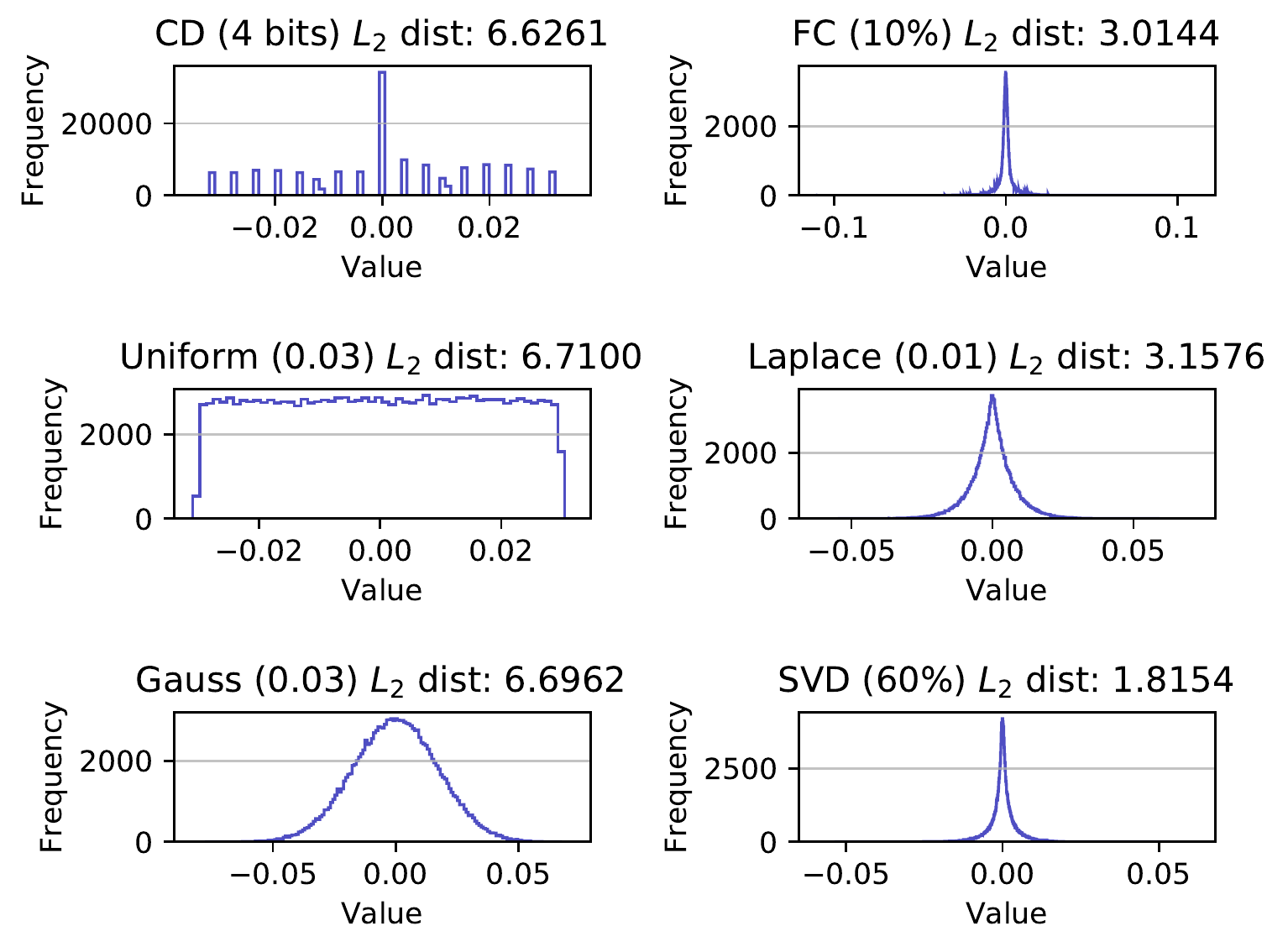}
     \caption{\label{fig:channel-pdf-histogram} Distribution of \textit{deltas} for input perturbation defences.}
\end{figure}

We emphasize in the paper that the distribution of noise does not matter as much as the magnitude. 
To illustrate this point, we plot the distribution of \textit{deltas} for six imprecise channels in Figure~\ref{fig:channel-pdf-histogram}. We compute the \textit{deltas} by subtracting an original image from the perturbed adversarial image and plot the histograms of differences. We use an image from the ImageNet dataset. For all the examples, the correct labels were recovered. We use the C\&W attack with 1000 iterations and the initial value $c=0.01$. 
All of the distributions are varied, yet they achieve similar robustness in the non-adaptive setting.


\section{Attack Input Perturbation Defenses Non-adaptively}
\label{subsection:NeighborhoodAdversarialExamples}

\begin{figure}[htb!]
    \centering
    \includegraphics[width=0.58\columnwidth]{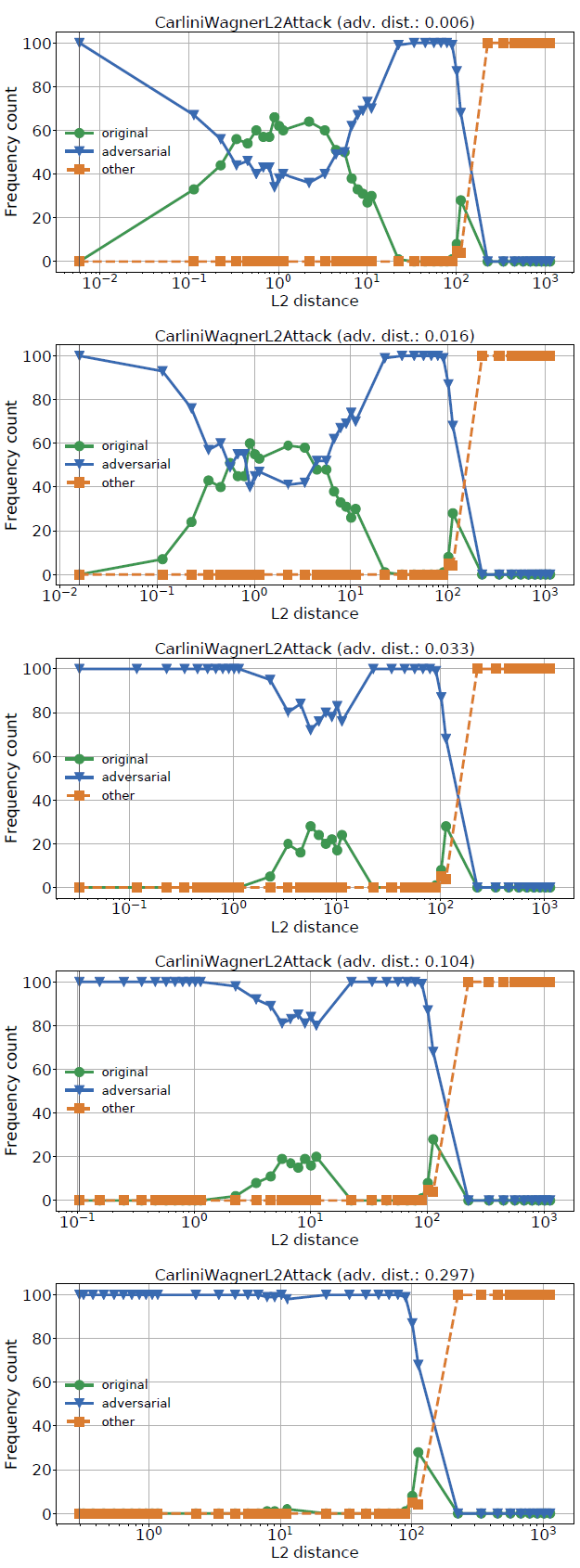}
    \caption{\label{fig:full_adv_neighboor}Frequency of model predictions for original, adversarial, and other classes as we increase the attack strength.}
\end{figure}

We present experiments for the analysis described in  Section~\ref{sec:recoveryWindow}.

The attack vector against input perturbation defenses simply makes the adversarial perturbations large enough that the defender significantly hurts the accuracy of the model when trying to dominate the adversarially placed strategic perturbations.

This interplay is visualized in Figure~\ref{fig:full_adv_neighboor}. We use an example from the ImageNet dataset, the ResNet-50 architecture, and set the stochastic channel to the Gaussian noise. We start from an adversarial example generated with the Carlini \& Wagner (non-adaptive) $L_2$ attack (left-hand side of each subplot) and for consecutive subplots (in the top to bottom sequence), we increase the attack strength and incur higher distortion of the adversarial image from the original image. For a single plot, we increase the $L_2$ distance of the output from the stochastic channel to the adversarial example by increasing the Gaussian noise (controlled by its standard deviation $\sigma$ and with mean $\mu=0$). For $L_2$ distances incurred by different noise levels, we execute 100 predictions. We use the frequency count and report how many times the model predicts the original, adversarial, or other class. 

The adversarial class is induced by a non-targeted attack. The \textbf{other class} is neither original (correct) nor adversarial and caused by (too much) Gaussian perturbation used in the defense. In this experiment, we begin with an adversarial example and systematically add more Gaussian noise. The model’s predictions change from the adversarial class, through original, and to a random other class.

The plot shows what range of distances from the adversarial image reveals the correct class. For the adversarial examples that are very close to the original image (e.g. adversarial distance of 0.006 for the top figure), the window of recovery (which indicates which strengths of the random noise can recover the correct label) is relatively wide, however, as we increase the distance of the adversarial image from the original image (by increasing the strength of the attack in consecutive plots), the window shrinks and finally we are unable to recover the correct label.
Figure~\ref{fig:all} shows that these large distortion attacks are still imperceptible to the human eye. To avoid any sensitivity to input perturbation defenses, an attacker can eliminate this recovery window by generating a larger distortion attack. Besides, statistical indications of adversarial examples should also be eschewed~\cite{OddsAreOdd}.

\begin{figure}[ht]
    \centering
    \includegraphics[width=0.8\columnwidth]{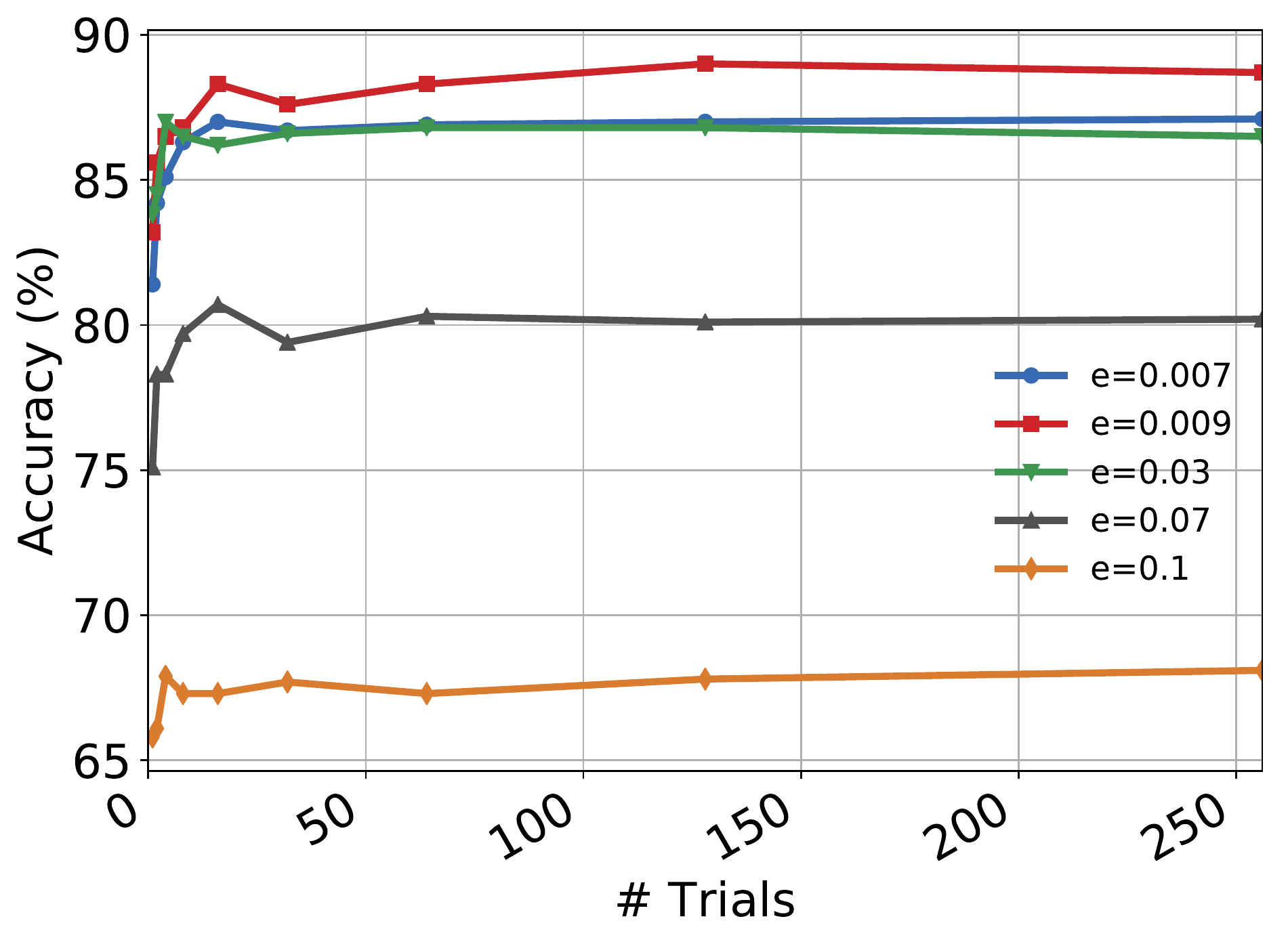}
        \caption{For the CIFAR-10 dataset, we run multiple trials of the uniform noise perturbation and take the most frequent prediction. We further test multiple noise levels. The multiple trials improve overall accuracy for different noise levels significantly. After 128 trials for the best setting we are within 3\% of the overall model accuracy (of about 93.5\%). \vspace{5em} \label{fig:trials}}
\end{figure}

\section{Transferability of the adversarial images}
\label{Sec:SupTransferability}
The details about the setup for experiment described in Section~\ref{sec:partially-adaptive-setting} and Table~\ref{table:attack-transfer}.

We use 30\% FC compression, 50\% SVD compression, 4-bit values in CD, 0.03 noise level for Gauss and Laplace, and 0.04 noise level for the Uniform channel. We use the ResNet-18 model, 2000 images from the CIFAR-10 test set, and 100 attack iterations with 5 binary steps to find the $c$ value (with initial $c$ value set to 0.01) for the adaptive C\&W $L_2$ attack.

We also run a similar experiment for the ImageNet dataset and present results in Table~\ref{table:attack-transfer-imagenet}.

\begin{table*}
\centering
\caption{Transferability of the adversarial images that extend results from Table~\ref{table:attack-transfer} for ImageNet dataset. We use 30\% FC compression, 50\% SVD compression, 4 bit values in CD, 0.03 noise level for Laplace, and 0.04 noise level for the Gauss and Uniform channels. We use 3000 images from the ImageNet validation set and 100 attack iterations.}
\label{table:attack-transfer-imagenet}
\begin{tabular}{ccccccc}
\toprule
\backslashbox{\textit{A}}{\textit{D}} & FC & CD & SVD & Gauss & Uniform & Laplace \\
\midrule
FC & 0.10 & 75.50 & 75.83 & 77.04 & 77.49 & 76.29 \\
CD & 0.17 & 1.16 & 6.77 & 62.60 & 62.04 & 65.46 \\
SVD & 12.02 & 72.33 & 0.46 & 72.79 & 72.52 & 73.09 \\
Gauss &    0.57 & 26.67 & 6.67 & 58.68 & 58.62 & 64.95 \\
Uniform & 0.50 & 26.71 & 6.99 & 58.48 & 59.06 & 64.59 \\
Laplace & 0.33 & 18.59 & 4.16 & 29.76 & 29.84 & 50.00 \\
\toprule
\end{tabular}
\end{table*}

\section{Multiple Trials For Stochastic Perturbations}
Another compelling reason to use stochastic perturbations (like Uniform noise) as a defense is that it can be run repeatedly in many random trials. We show that doing so slightly improves the efficacy of the defense.
We randomly choose 1000 images from the CIFAR-10 test set.
For each of these images, we generate an adversarial attack.
We then pass each image through the same stochastic perturbation multiple times.
We take the most frequent prediction. 
Figure~\ref{fig:trials} illustrates the results.

Only 16 trials are needed to get a relatively strong defense. We argue that this result is significant. Randomized defenses are difficult to attack. The attacker cannot anticipate which particular perturbation to the model will happen. The downside is the potential of erratic predictions. We show that a relatively small number of trials can greatly reduce this noise. Furthermore, the expense of running multiple trials of a randomized defense is small relative to the expense of synthesizing an attack in the first place.

\begin{figure}[ht]
    \includegraphics[width=1.0\columnwidth]{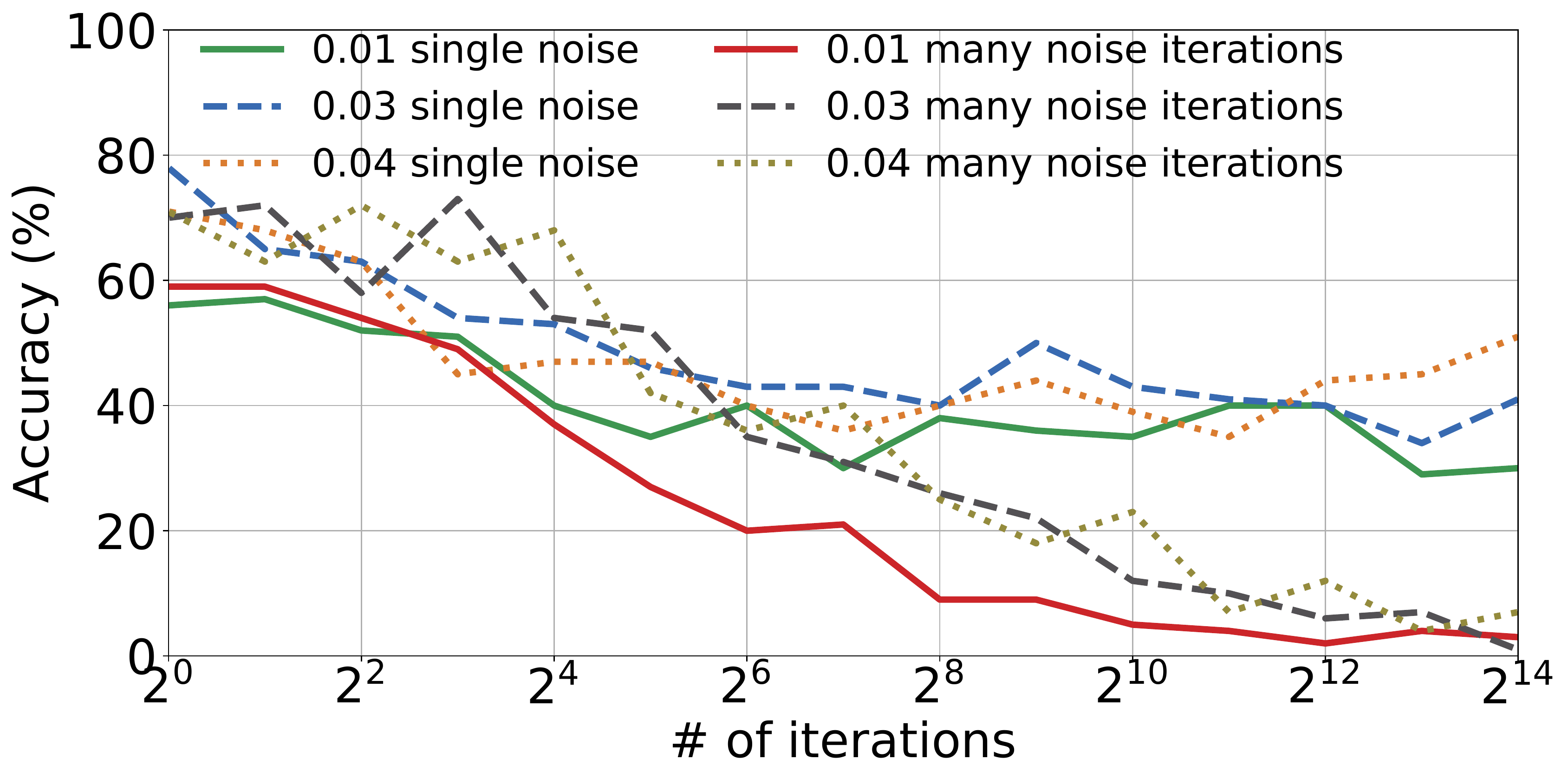}
        \caption{For the CIFAR-10 dataset, we run multiple trials of the uniform noise perturbation and take the most frequent prediction in the defense (many noise iterations). We also run just a single noise injection and return the predicated label. The attacker runs the same number of many uniform trials as the defender. The experiment is run on 100 images, with 100 C\&W $L_2$ attack iterations.  \label{fig:manyIters}}
\end{figure}

\section{More Details on The White-Box Adaptive Attack}
\label{white-box-adaptive-attack-sup}

\begin{figure}
    \centering
    \includegraphics[width=0.7\columnwidth]{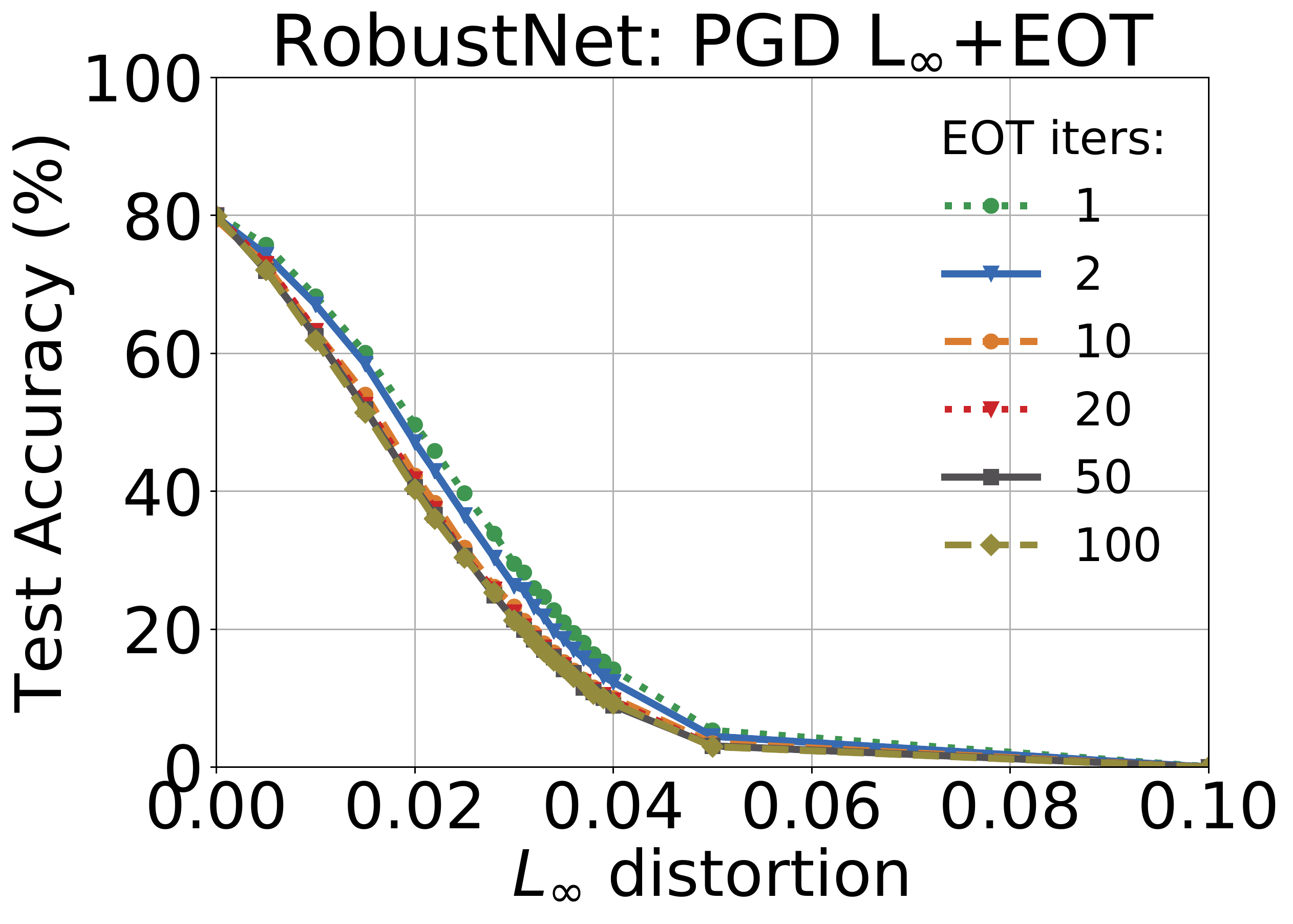}
    \caption{
    \emph{\textbf{PGD + EOT.}}
    We use the PGD attack with 40 iterations and EOT with different number of iterations tested against RobustNet. We train ResNet-20 model on CIFAR-10 dataset. Clean accuracy is 88\%.
    \label{fig:pgd-eot}}
\end{figure}

We also test RobusetNet against the adaptive PGD + EOT attack. We present the results in Figure~\ref{fig:pgd-eot}.

\begin{figure}
    \centering
    \includegraphics[width=0.7\columnwidth]{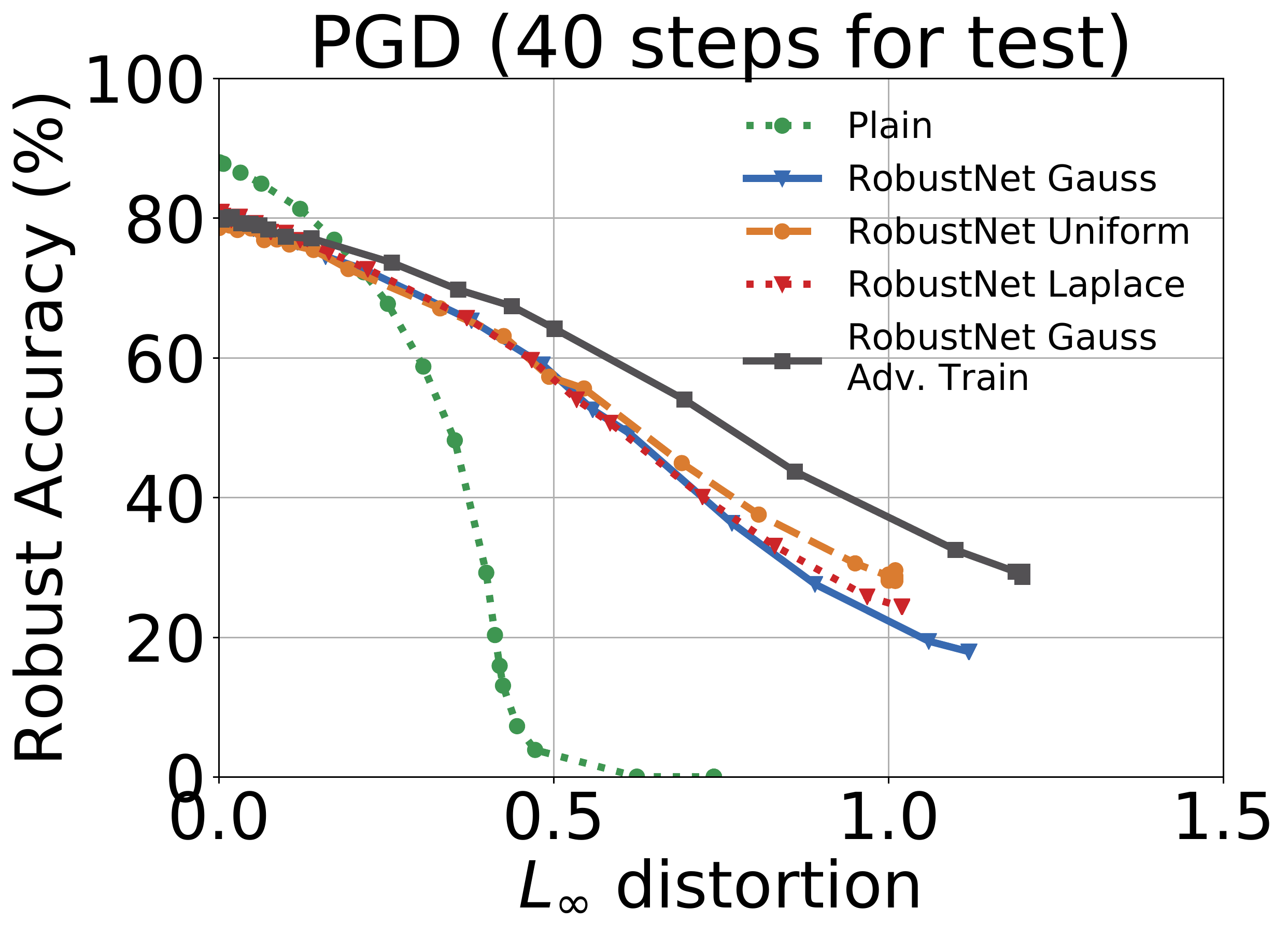}
    \caption{
    \emph{\textbf{Different types of noise for RobustNet.}}
    We use the PGD attack with 40 iterations. We train ResNet-20 model on CIFAR-10 dataset. Clean accuracy is 88\%. The Gauss, Uniform, and Laplace noise provide the same robustness.
    \label{fig:robustnet-gauss-uniform-laplace}}
\end{figure}

We test different types of noise injected into RobustNet and present results in Figure~\ref{fig:robustnet-gauss-uniform-laplace}.

The stochastic perturbations are harder to attack with gradient-based adaptive methods.
Our strategy is to send output from the adversarial algorithm through the imprecise network and the network at least as many times as set in the defense. We mark the attack as successful if the most frequent output label is different from the ground truth. The more passes through the noisy channel we optimize for, the stronger the attack. Furthermore, we run many iterations of the attack to decrease the $L_{2}$ distortion. An attack that always evades the noise injection defense is more difficult to generate because of randomization. The other randomized approach was introduced in dropout~\cite{feinman2017detecting}. The attacks against randomized defenses require optimization of complex loss functions, incur higher distortion, and the attacks are not fully successful~\cite{carlini2017adversarial}.
In the Figure~\ref{fig:manyIters}, we present the result of running attacks and defenses on CIFAR-10 data with single and many iterations. The defense with many trials can be drawn to 0\% accuracy, however, the defense not fully optimized by the adversary (single noise injection) can result in about 40\% or higher accuracy.

\section{Gradient-based Analysis}
\label{appendix:subsection:gradientBasedAnalysis}
\begin{figure}[ht]
    \centering
    \includegraphics[width=1.0\linewidth]{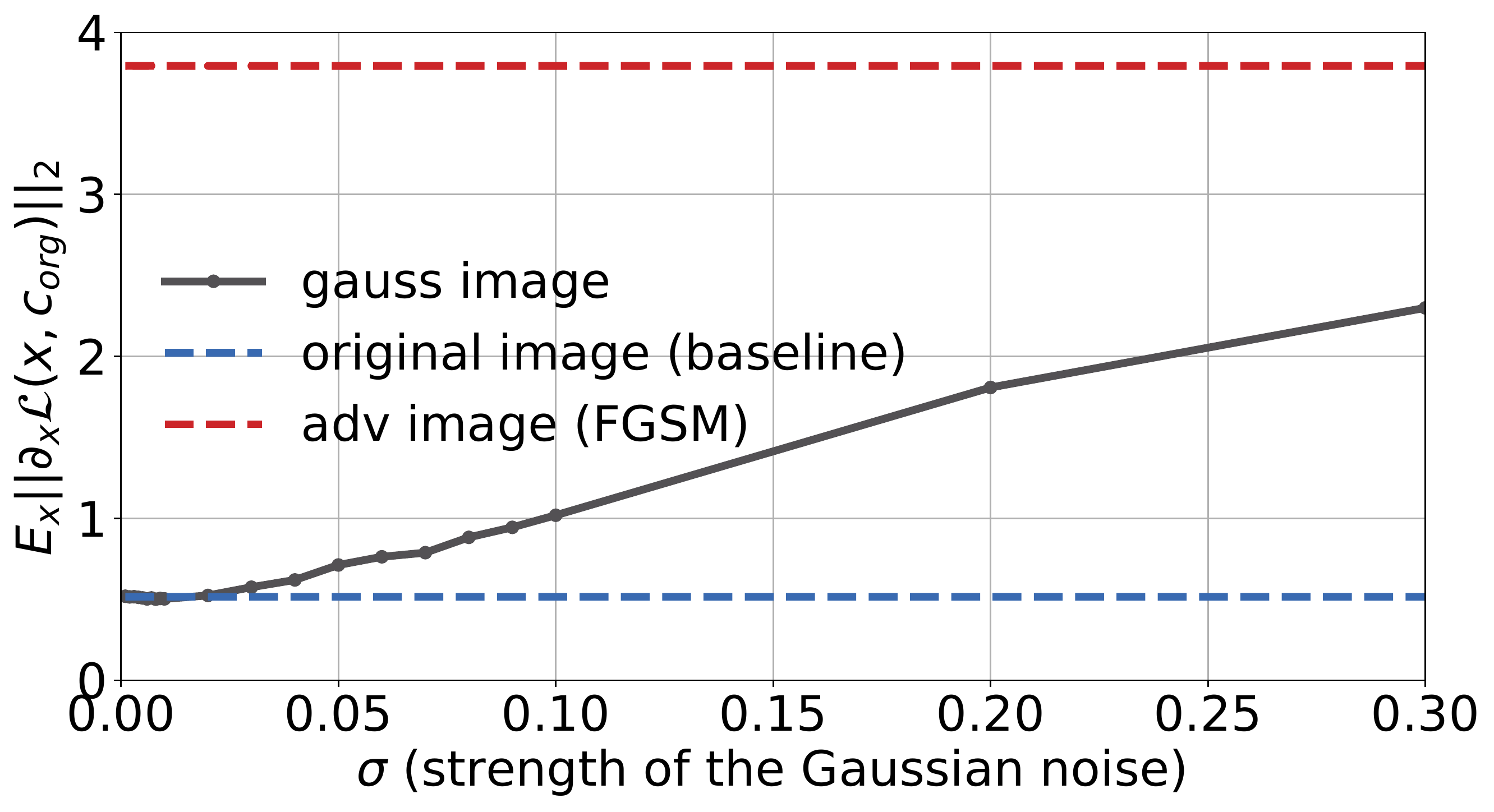}
     \caption{\label{fig:gauss_grads} The changes in the $L_2$ norm of the gradient of the loss w.r.t. the input image $x$ for the correct class $c_{org}$ as we add Gaussian noise to the original image. The experiment is run on 1000 images from the ImageNet dataset.}
\end{figure}

\begin{figure}[ht]
    \centering
    \includegraphics[width=1.0\linewidth]{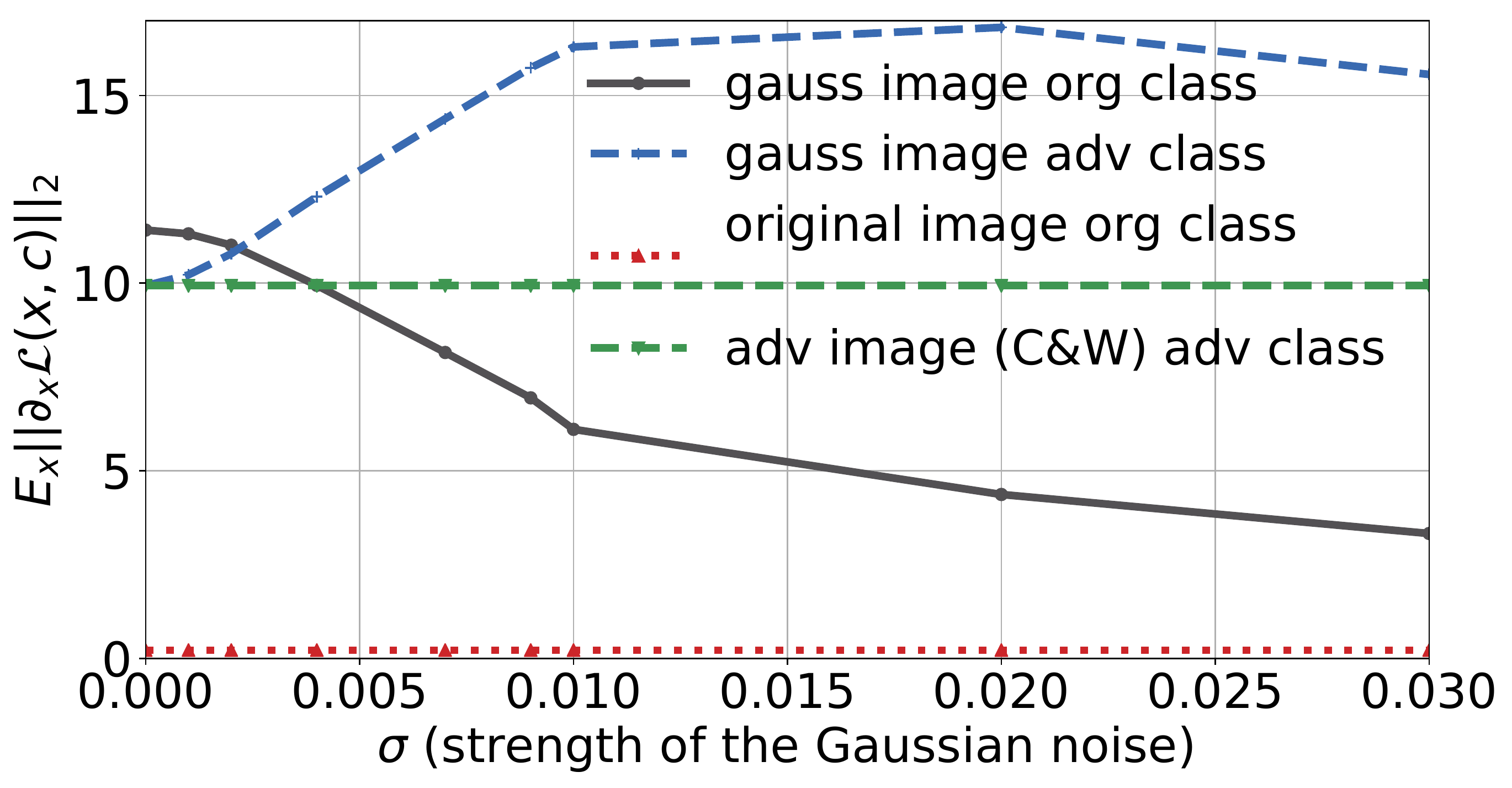}
     \caption{\label{fig:gauss_grads_adv} The changes in the $L_2$ norm of the gradient for the correct class $c_{org}$ and the adversarial class $c_{adv}$ as we add Gaussian noise to the adversarial image generated with C\&W $L_2$ attack. The experiment is run on 1000 images from the CIFAR-10 dataset.}
\end{figure}

\begin{figure}[ht]
    \centering
    \includegraphics[width=1.0\linewidth]{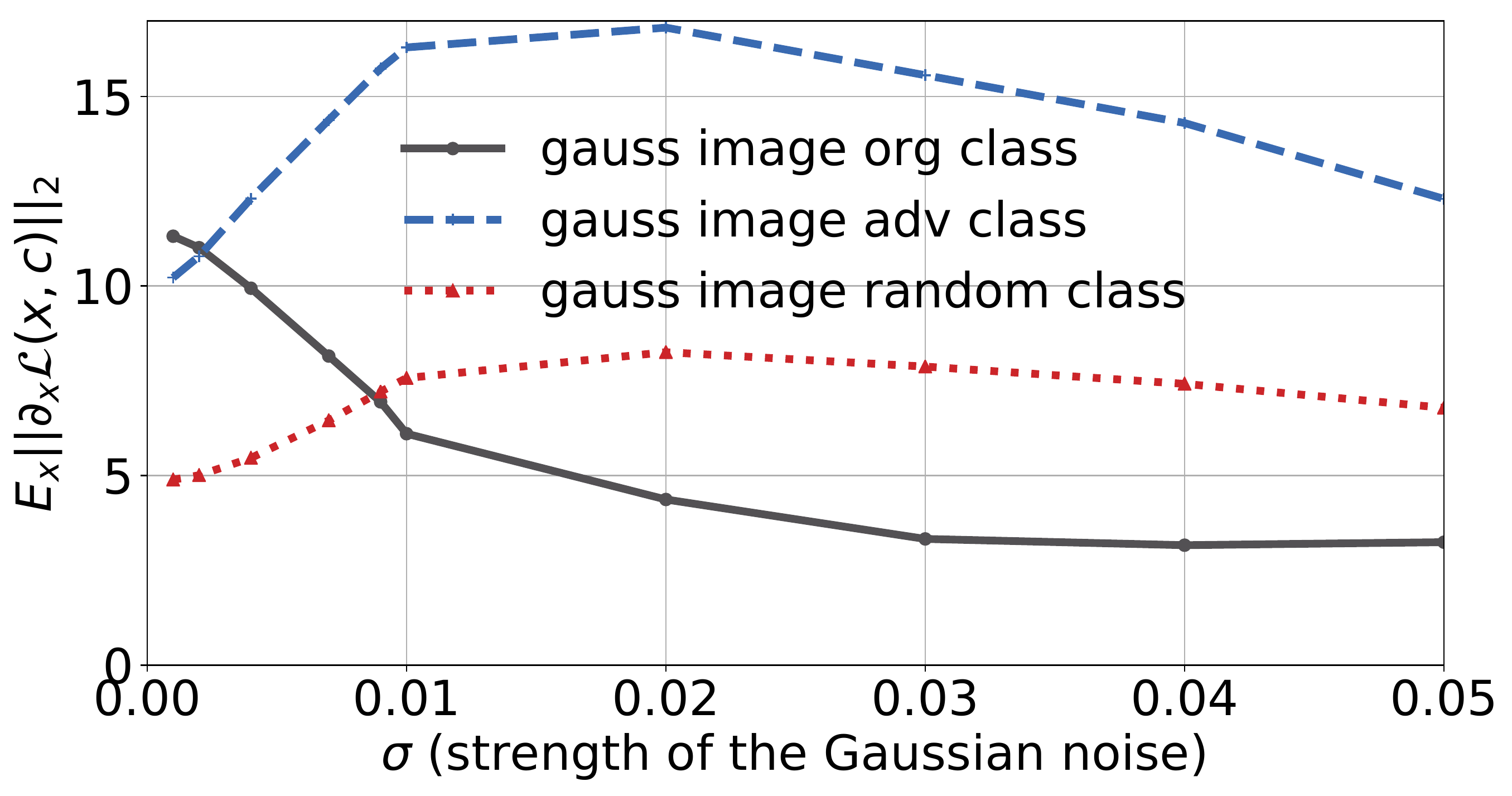}
     \caption{\label{fig:adv_gauss_grads_random} The changes in the $L_2$ norm of the gradient of the loss for the correct class $c_{org}$, the adversarial class $c_{adv}$, and a random class $c_{ran}$ as we add Gaussian noise to the adversarial image generated with C\&W $L_2$ attack. The experiment is run on 1000 images from the CIFAR-10 dataset.}
\end{figure}

We start the analysis from clean images that are classified correctly. We present how the gradient of the loss w.r.t. the input image changes for the correct class as we add the Gaussian noise to the original image in Figure~\ref{fig:gauss_grads}. The norm of the gradient smoothly increases. 
In Figure~\ref{fig:gauss_grads_adv}, we start from an adversarial image found with the default C\&W attack from the foolbox library. Then, we systematically add Gaussian noise to the adversarial image and collect data on the norm of gradients for the original and adversarial classes. The norm of the gradients for the adversarial class increases while the norm of the gradients for the original class decreases. We cross the decision boundary to the correct class very early and recover the correct labels for images. Then, as we add substantially more Gaussian noise, the predictions of the classifier become random and the norms of the gradients converge to a single value. In Figure~\ref{fig:adv_gauss_grads_random}, we plot the gradients also for a random class. We observe that for an untargeted attack, the gradients for the original and adversarial classes are larger than for the other classes. The targeted attack decreases the loss for the target class and the gradients for the adversarial classes are lower when compared with gradients from the untargeted attacks, so fewer images can be recovered in the former case. The targeted attack causes a smaller increase in the norms of gradients for the original class than the untargeted attack. However, it is still higher than for a random class.

\begin{figure}[ht]
    \centering
    \includegraphics[width=1.0\linewidth]{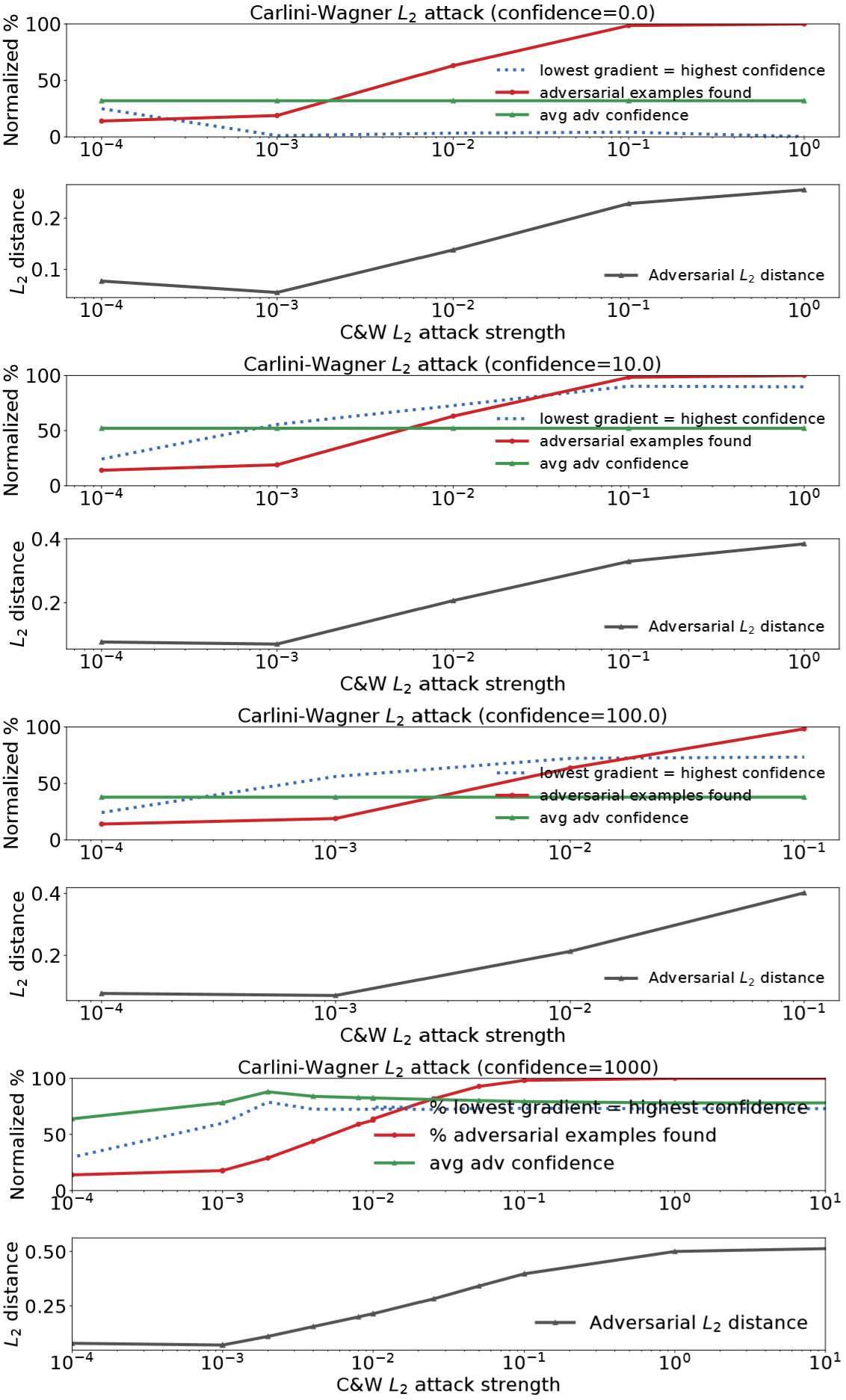}
    \caption{\label{fig:attack_conf_cw} Dependence between confidence levels of the CW attack and gradients of the generated adversarial examples.}
\end{figure}

\subsection{Controlling the Gradient Magnitude}
The gradient magnitude can be controlled with the confidence parameter in the C\&W attack. There is a high dependence between what confidence level is set for the attack and what the gradient norm is. If we set the confidence to 0 (zero), then we only intend to misclassify an example, and the gradient norms are very low (also the prediction confidence of the misclassified examples is low). The gradient norm is w.r.t. an adversarial image for its adversarial class. If we set the confidence of the attack to a higher level, then also the gradient is smaller for the adversarial examples and the prediction confidence (for adversarial classes) is increased. The results are not consistent as we would expect the highest prediction confidence for the highest attack confidence. We find that overall the attack confidence for the C\&W attack of about 1000 gives us the highest average confidence of predictions (from the model that the given adversarial example is of the adversarial class). However, for the attack confidence level of about 10, we can reach almost 100\% prediction confidence for very high strength of the attack.
For the consecutive plots in Figure~\ref{fig:attack_conf_cw} we increase the confidence level for the C\&W attack. In each plot, we average confidence in predictions across images as we systematically increase the attack strength. For the attack confidence levels 0, 10, 100, we run the experiments on 1000 images from the CIFAR-10 test set. For the attack confidence level 1000, we run the experiment on the whole CIFAR-10 test set.

The softmax probabilities and norms of gradients are correlated for original images but not necessarily for adversarial examples. We run the experiment for 1000 images from the CIFAR-10 dataset. The classification accuracy on the clean data is 93.9\%. Next, we take only the correctly classified images and for each image, we generate an adversarial example using the default C\&W attack from the foolbox library (where the confidence parameter is set to zero). We record the softmax probabilities and norms of the gradients for each of the 10 classes. For 99\% of the original images, the lowest gradients are for the original class. Only for 1\% of the adversarial examples, the lowest gradients are for the adversarial class.



\section{Hessian-based Analysis}
\begin{figure}[t]
     \centering
    \includegraphics[width=1.0\linewidth]{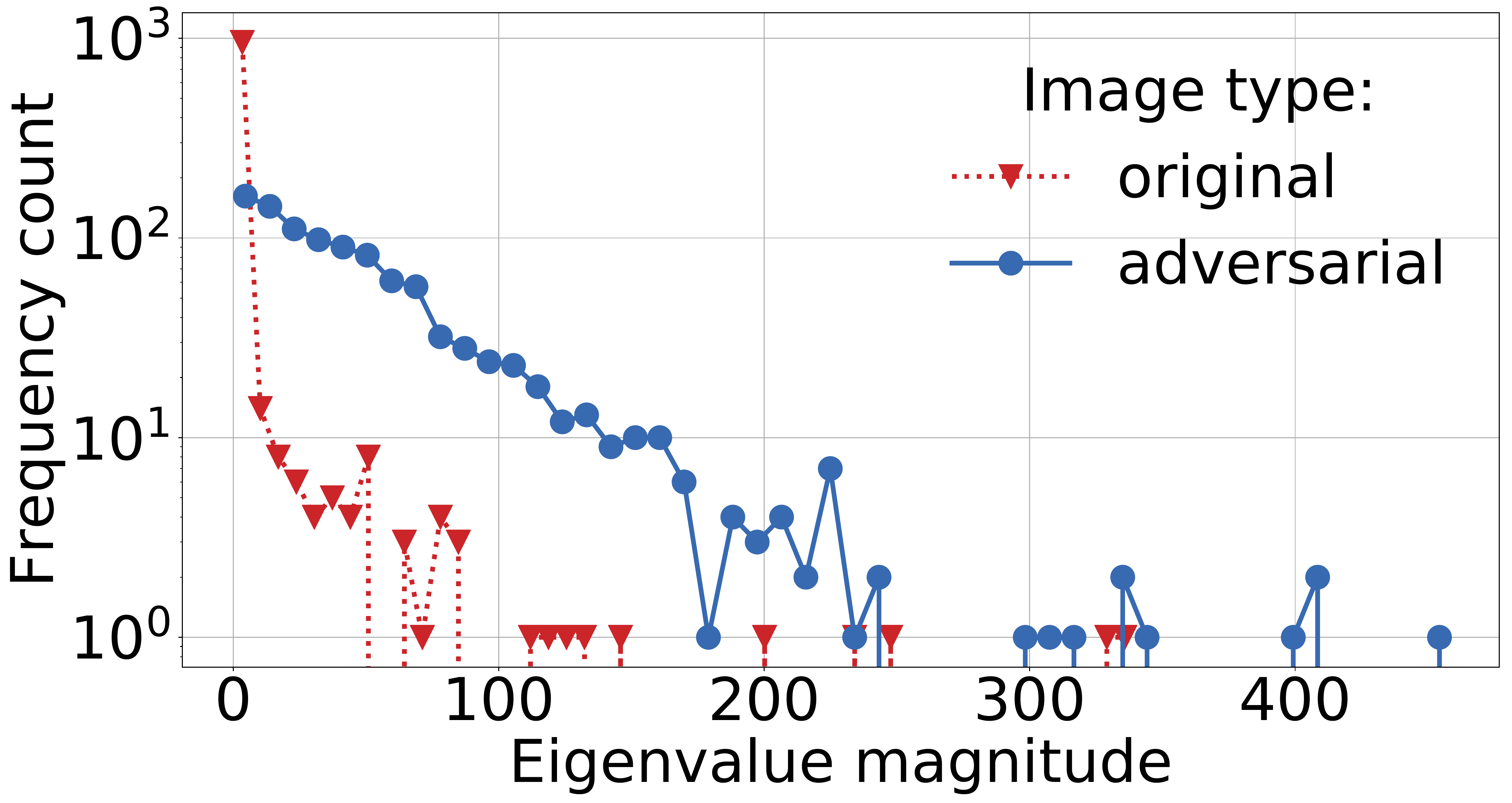}
    \caption{\label{fig:hessian-cifar10} Histogram of top eigenvalues of the Hessians w.r.t. the input 1024 images from the CIFAR-10 dataset trained on the ResNet-18 architecture.}
\end{figure}

We present the Hessian spectrum in Figure~\ref{fig:hessian-cifar10}. For the adversarial images, the eigenvalues are higher, which indicates higher instability and proclivity to prediction changes as defined in Section~\ref{subsection:perturbationAnalysis}.

\section{Visualizations of attacks and imprecise channels}
\begin{figure*}[b!]
    \centering
    \includegraphics[width=\textwidth]{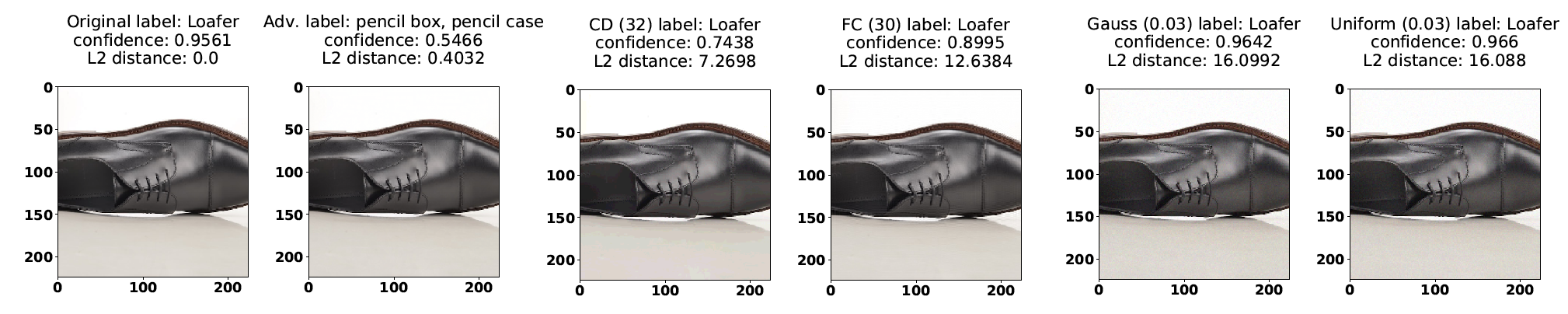}
     \caption{\label{fig:all} We plot a sample image from the ImageNet dataset in its original state, after adversarial (white-box, non-adaptive C\&W $L_2$) attack, and then after recovery via imprecise channels: CD (color depth reduction with 32 bits), FC (30\% compression in the frequency domain), Gaussian, and uniform noise ($\epsilon=0.03$).}
\end{figure*}

Figure~\ref{fig:all} presents a sample image from ImageNet for the Carlini-Wagner L2 attack.

\section{Regularization}
Perturbing the inputs and internal feature maps during training is equivalent to a form of Lipschitz regularization. However, defenses studied in the paper add noise also during inference and are not standard forms of regularization (e.g. Tikhonov or Lasso). We do show that training the models with the anticipated noise (that is added as a defense at inference time) is a valuable optimization but we do not believe that this is fully a regularization effect. The techniques are also akin to data augmentation. Furthermore, defenses such as feature squeezing are non-differentiable and do not fit into a standard regularization framework. Finally, regularization methods have to be designed and tuned specifically to create a strong defense.

\section{Additional Experiments for black-box attacks}
As a black-box attack, we define an attack that does not need knowledge about the gradient or the model.

\subsection{Decision-based attacks}
The attacks require neither gradients nor probabilities. They operate directly on the images.

\subsubsection{Robustness to Uniform and Gaussian Noise}


We evaluate the robustness of band-limited CNNs. Specifically, models trained with more compression discard part of the noise by removing the high-frequency Fourier coefficients (FC channel). In Figure~\ref{fig:all-attacks}, we show the test accuracy for input images perturbed with different levels of uniform and Gaussian noise, which is controlled systematically by the sigma parameter, fed into models trained with different compression levels (i.e., 0\%, 50\%, or 85\%) and methods (i.e., band-limited vs. RPA-based\footnote{The Reduced Precision Arithmetic, where operations on 16-bit floats are used instead of on 32 or 64-bit float numbers.}). Our results demonstrate that models trained with higher compression are more robust to the inserted noise. Interestingly, band-limited CNNs also outperform the RPA-based method and under-fitted models (e.g., via early stopping), which do not exhibit the robustness to noise.

Input test images are perturbed with uniform or Gaussian noise, where the sigma parameter is changed from 0 to 1 or 0 to 2, respectively. The more band-limited model, the more robust it is to the introduced noise.

\subsubsection{Contrast Reduction Attack}
This black-box attack gradually distorts all the pixels:
\begin{align*}
  \text{target} &= \frac{\text{max} + \text{min}}{2} \\
  \text{perturbed} &= (1 - \epsilon) * \text{image} + \epsilon * \text{target}
\end{align*}
where min and max values are computed across all pixels of images in the dataset.

We can defend the attack with CD (Color Depth reduction) until a certain value of epsilon, but then every pixel is perturbed smoothly so there are no high-frequency coefficients increased in the FFT domain of the image. The contrast reduction attack becomes a low-frequency based attack when considered in the frequency domain. Another way to defend the attack is to run a high-pass filter in the frequency domain instead of the low-pass filter.

We run the experiments for different models with CD and two band-limited models (the model with full spectra and no compression as well as a model with 85\% of compression - with FC layers). The CD does defend the attack to some extent and the fewer pixels per channel (the \textit{stronger} the CD in a model), the more robust the model is against the contrast reduction attack.

Test accuracy as a function of the contrast reduction attack for ResNet-18 on CIFAR-10 (after 350 epochs) is plotted in Figure~\ref{fig:all-attacks}. We control the strength of the attack with parameter epsilon that is changed systematically from 0.0 to 1.0. We use the whole test set for CIFAR-10. R denotes the number of values used per channel (e.g., R=32 means that we use 32 values instead of standard 256).


\subsubsection{Multiple pixels attack}
The foolbox library supports a single-pixel attack, where a given pixel is set to white or black. A certain number of pixels (e.g., 1000) is chosen and each of them is checked separately if it can lead to the misclassification of the image. The natural extension is to increase the number of pixels to be perturbed, in a case where the single-pixel attack does not succeed. We present results for the multiple pixel attack in Figure~\ref{fig:all-attacks}.

\subsection{Spatial-based attacks}
Spatial attacks apply adversarial rotations and translations that can be easily added to the data augmentation during training. However, these attacks are defended neither by removing the high-frequency coefficients nor by quantization (CD)
. We separately apply rotation by changing its angle from 0 to 20 degrees and do the translations within a horizontal and vertical limit of shifted pixels (Figure~\ref{fig:all-attacks}). 

\subsection{Score-based attacks}
The score based attack requires access to the model predictions and its probabilities (the inputs to the softmax) or the logits to estimate the gradients. 

\subsubsection{Local Search Attack}
\begin{figure*}[h!]
\centering
  \includegraphics[width=0.7\linewidth]{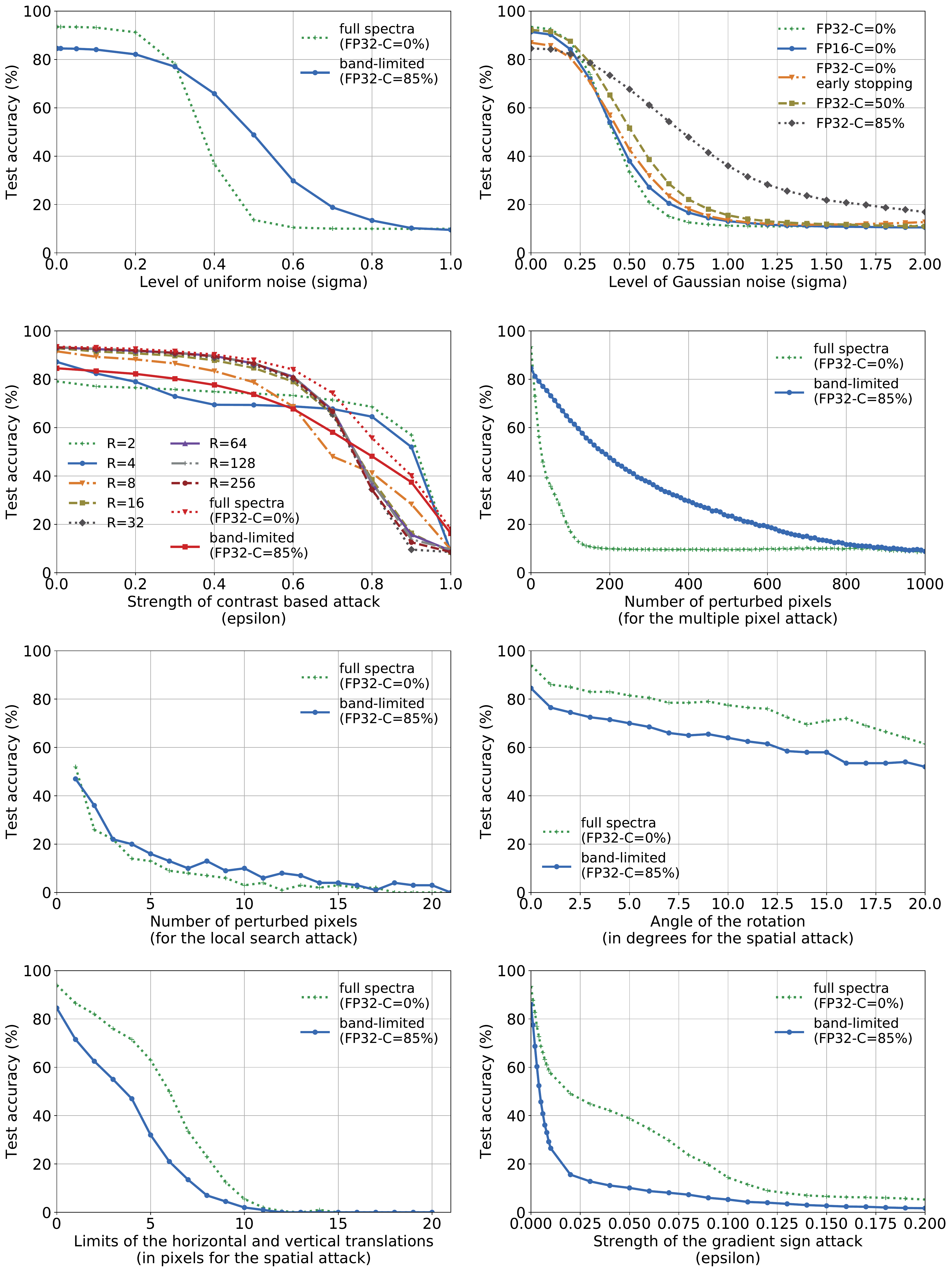} 
  \caption{Test accuracy as a function of the strenghts of the attacks for ResNet-18 on CIFAR-10.}
  \label{fig:all-attacks}
\end{figure*}

The local search attack estimates the sensitivity of individual pixels by applying extreme perturbations and observing the effect on the probability of the correct class. Next, it perturbs the pixels to which the model is most sensitive. The procedure is repeated until the image is misclassified, searching for additional critical pixels in the neighborhood of previously found ones. We run the experiments for the attack on 100 test images from CIFAR-10 since the attack is relatively slow (Figure ~\ref{fig:all-attacks}).

\section{Other domains}
We replicated our experiments on time-series data from the UCR archive and results suggest a similar relationship between perturbation and adversarial robustness. On an ECG dataset using the CW $L_2$-norm attack, we observe that: (1) there is similarly a narrow recovery window of $L_2$ distortions ranging from 0.7 to 1.6 where the perturbation is a successful defense, and (2) layer-perturbation methods are the most effective adaptive defense.

We use NonInvasiveFetalECGThorax1 dataset. A 3-layer RobustNet based on FCN architecture for time-series data is set with noise levels 0.06, 0.05, 0.05. The results show that combining adversarial training with RobustNet gives superior performance. We used Adv. Training with strength 0.04 (i.e., max $L_{\infty}$ norm for 7 step PGD attack). 


\end{document}